\documentclass[letterpaper]{article} 
\usepackage[submission]{aaai24}  
\usepackage{times}  
\usepackage{helvet}  
\usepackage{courier}  
\usepackage[hyphens]{url}  
\usepackage{graphicx} 
\usepackage{natbib}  
\usepackage{caption} 
\usepackage{algorithm}
\usepackage{algorithmic}

\usepackage{enumitem}
\usepackage{latexsym}
\usepackage{multirow}
\usepackage{adjustbox}
\usepackage{makecell}
\usepackage{scrextend}
\usepackage{amsmath}

\usepackage{framed}
\usepackage[table,xcdraw]{xcolor}

\usepackage{newfloat}
\usepackage{listings}
\DeclareCaptionStyle{ruled}{labelfont=normalfont,labelsep=colon,strut=off} 
\floatstyle{ruled}
\newfloat{listing}{tb}{lst}{}
\floatname{listing}{Listing}
\title{Large Language Models as Financial Data Annotators: A Study on Effectiveness and Efficiency}
\author{
    Toyin Aguda\equalcontrib,
    Suchetha Siddagangappa\equalcontrib,
    Elena Kochkina,
    Simerjot Kaur,\\
    Dongsheng Wang,
    Charese Smiley,
    Sameena Shah
    \\
  JPMorgan AI Research \\
  \texttt{\{toyin.d.aguda, suchetha.siddagangappa, elena.kochkina, simerjot.kaur, dongsheng.wang,charese.h.smiley, sameena.shah\}@jpmchase.com}
}
\affiliations{
    \textsuperscript{\rm 1}JPMorgan AIResearch\\

}

\usepackage{bibentry}

\begin{document}

\maketitle

\begin{abstract}

Collecting labeled datasets in finance is challenging due to scarcity of domain experts and higher cost of employing them. While Large Language Models (LLMs) have demonstrated remarkable performance in data annotation tasks on general domain datasets, their effectiveness on domain specific datasets remains underexplored. To address this gap, we investigate the potential of LLMs as efficient data annotators for extracting relations in financial documents. 
We compare the annotations produced by three LLMs (GPT-4, PaLM 2, and MPT Instruct) against expert annotators and crowdworkers.  We demonstrate that the current state-of-the-art LLMs can be sufficient alternatives to non-expert crowdworkers. We analyze models using various
prompts and parameter settings and find that customizing the
prompts for each relation group by providing specific examples belonging to those groups is paramount. 
Furthermore, we introduce a reliability index (LLM-RelIndex) used to identify outputs that may require expert attention.
Finally, we perform an extensive time, cost and error analysis and provide recommendations for the collection and usage of automated annotations in domain-specific settings. 

\end{abstract}

\section{Introduction}
Financial NLP (FinNLP) is an active and growing research area with numerous applications in analyzing and comprehending financial texts. The development of effective FinNLP models relies on well-annotated datasets derived from financial documents. However, annotating such datasets is challenging as it requires a deep understanding of financial concepts to decipher the complex terminologies and calculations present in the documents. Crowdsourcing platforms are generally used for annotations. While they are generally effective for tasks that do not require high levels of expertise, they often produce inconsistent and inaccurate annotations when it comes to domain-specific datasets. This approach requires careful instruction crafting, multiple annotation rounds, increased number of workers, and, finally, expert intervention for enhanced accuracy and consistency. 

\begin{figure}[h!]
\footnotesize
 \begin{framed}
 \begin{flushleft}
\textbf{Text}: The predecessor \underline{\textcolor{red}{Mississippi Power Company}} was incorporated under the laws of the State of Maine on November 24, 1924 and was admitted to do business in Mississippi on  \underline{\textcolor{blue}{December 23, 1924}} and in Alabama on December 7, 1962.\\
 \end{flushleft}
\vspace{0.1cm}
\hrule
\vspace{0.2cm}
\textbf{Relation type:} Organization--Date \\ 
\vspace{-0.2cm}
\hrule
\vspace{0.2cm}
\textbf{Expert Label:} \textsc{No/other relation} \\
\textbf{Crowdworker Label:} \textsc{Formed on}
 
 \end{framed}
 \caption{Example of relation extraction task from REFinD dataset.}
 \label{fig:example}
\end{figure}

The wide array of tasks in which Large Language Models (LLMs), such as GPTs~\cite{10.5555/3495724.3495883,OpenAI2023GPT4TR}, have demonstrated state-of-the-art zero-shot capabilities naturally raises the question of whether these models have the potential to substitute for human annotators. Using LLMs as data annotators can offer a lot of advantages such as cost-effectiveness, scalability and potential for iterative improvement. However, strong performance on benchmark datasets alone does not ensure a model's suitability to replace human annotators. In addition to accuracy, consistency and biases associated with this approach needs to be carefully studied. 

While positive results of using LLMs as annotators for general-domain tasks have been reported in recent papers and preprints~\cite{doi:10.1073/pnas.2305016120,tornberg2023chatgpt}, their performance in specialized domains such as finance remains underexplored. In this work, we assess the efficacy of LLMs as data annotators for financial relation extraction task using REFinD dataset~\cite{10.1145/3477495.3532019}. 
\def\thefootnote{*}\footnotetext{These authors contributed equally to this work}\def\thefootnote{\arabic{footnote}}

The relation extraction task in financial documents involves identifying specific relations between financial entities such as companies and persons. Financial relation extraction presents unique challenges due to the domain-specific nature of financial language and the scarcity of labeled data. General relation extraction models trained on generic tasks may lack the necessary understanding of finance-specific terms, leading to difficulties in capturing nuanced patterns. For example, certain relations, such as board membership versus employment, require domain expertise for accurate interpretation. Ambiguity further complicates the task, as implicit relationships, like company acquisitions based on stock ownership, may be challenging for generic models to identify. Furthermore, financial sentences are notably more complex, with longer average lengths and greater entity pair distances compared to generic domains, as demonstrated in REFinD.

Figure \ref{fig:example} shows an example from the REFinD dataset where we are interested in finding a relation between an \textsc{organization-date}
entity pair, wherein we are interested in extracting the relation between
an organization - \textit{Mississippi Power Company} and date - \textit{December 23, 1924}.  For this entity pair, the relation label options presented to experts and crowdworkers are (i) \textsc{formed on} (ii) \textsc{acquired on} and (iii) \textsc{no/other relations}. The label chosen by experts is \textsc{no/other relation}, the reason being Mississippi Power Company was formed on November 24, 1924 and not on December 23, 1924. However, crowdworkers incorrectly identified the label as \textsc{formed on}. This discrepancy between expert labels and crowdworker labels highlights the difficulty of financial relation extraction tasks.

In this work, we compare the output of LLMs and crowdworkers against expert annotations, extending our analysis beyond performance metrics and addressing time, cost and reliability aspects of the annotation process. Our contributions are the following:
(i) To the best of our understanding, we are the first in the financial domain to demonstrate the capabilities of LLMs as data annotation tools by evaluating them against domain experts and crowdworkers. 
(ii) We compare 3 models (GPT-4, PaLM 2, and MPT Instruct) and parameters (varying temperature, random seed and prompting approaches) to identify the most accurate and reliable configuration. (iii) We introduce reliability index, a metric designed to identify trustworthy samples and filter out those requiring human intervention. 
(iv) We demonstrate that LLMs can replace non-expert crowdworkers for a significant portion of the dataset, while expert intervention is necessary for the remaining instances to ensure accurate annotations. We also offer guidance on best practices for implementing LLMs in the annotation process.

\section{Related Work}
\label{sec:related_work}

\citet{wang-etal-2021-want-reduce} pioneered the use of GPT-3~\cite{10.5555/3495724.3495883} as a cost-effective data labeler for training models. The potential of LLMs as data annotators has been explored in various tasks including relevance, stance, topic and frame classification~\cite{doi:10.1073/pnas.2305016120}, sentiment analysis, hate speech detection~\cite{zhu2023can,10.1145/3543873.3587368}, political affiliation~\cite{tornberg2023chatgpt} and news classification~\cite{reiss2023testing}\footnote{Note that some of the citations are recent publicly available preprints.}.  Since the majority of these tasks do not require the domain expertise of a human annotator, the effectiveness of LLMs in domain-specific datasets remains underexplored. This study investigates LLMs' potential in the financial domain. 

Existing literature on the application of LLMs in the financial domain remains sparse. \citet{li-etal-2023-chatgpt} have evaluated the performance of GPT-3.5 and GPT-4 on various finance benchmark datasets and reported strong performance on arithmetic reasoning, news classification and financial named entity recognition. However, this study did not consider the potential use of LLMs as annotators in comparison to non-expert crowdworkers, or the relation extraction task, which is the focus of our paper. 

Several approaches assess the potential of LLMs as data annotators. Studies like \citet{kuzman2023chatgpt,chiang-lee-2023-large,ding-etal-2023-gpt} explore different aspects of LLMs including  comparing zero-shot performance of ChatGPT against a task-specific fine-tuned model, and measuring the alignment of LLM and human evaluations. \citet{he2023annollm,tornberg2023chatgpt,doi:10.1073/pnas.2305016120} compare the model outcomes with crowdworkers and expert annotators. While the latter approach is more costly, we adopt it in this study due to its direct relevance to our research question.

LLMs as annotators yield mixed results, with some studies showing higher performance than humans~\cite{doi:10.1073/pnas.2305016120,tornberg2023chatgpt}, while others highlight limitations in new domains~\citet{zhu2023can} and consistency issues~\cite{reiss2023testing}. \citet{zhu2023can}  report GPT's overestimation of certain classes. This further motivates our study to evaluate these aspects for finance domain specifically. 

It is also worth noting that  most studies focus on GPT models only ~\cite{10.1145/3543873.3587368,reiss2023testing,tornberg2023chatgpt,zhu2023can,ding-etal-2023-gpt}. We address this limitation by comparing three generative LLMs, GPT-4~\citep{OpenAI2023GPT4TR}, PaLM 2~\citep{anil2023Palm}, MPT Instruct~\citep{MosaicML2023Introducing}, each with different size, training data, and procedures. 

\section{Dataset} 
Our experiments utilize the REFinD dataset~\cite{10.1145/3477495.3532019}. Derived from texts within quarterly and annual reports of publicly traded companies (10-X), REFinD is the largest dataset available for financial relation extraction. This is also the only financial domain dataset for which we were able to obtain annotations broken down into expert and individual crowdworkers.
REFinD dataset has 28,676 instances and 22 relations types across 8 entity pairs. The only other available Financial relation extraction dataset FinRED~\citep{sharma2022finred} is significantly smaller (6,767 instances and 29 relation types) and does not release annotations provided by individual crowdworkers. 


These 8 entity pairs covered in REFinD include \textsc{person--title}, \textsc{person--organization}, \textsc{person--university}, \textsc{person--government agency}, \textsc{organization--gpe}, \textsc{organization--date}, \textsc{organization--organization} and \textsc{organization--money}. Each entity pair includes several finance-oriented relation types. The choice of this dataset is further justified by the fact that it was released in mid 2023, which makes it unlikely to have been part of the training data for the selected LLMs. For our experiments, we utilize 3598 instances from the test set of REFinD, due to the costs associated with LLMs usage.


\section{Experiments}
\label{sec:exp}
In this section, we present comprehensive descriptions of the generative models, prompts, and evaluation metrics utilized in our study. 

\subsection{Models}
In our experiments, we employed three Large Language Models (LLMs), GPT-4, PaLM 2, and MPT Instruct, selected based on their exceptional performance in benchmark leaderboards\footnote{\url{https://huggingface.co/spaces/HuggingFaceH4/open_llm_leaderboard}}, accessibility, API availability, and permissive licenses. These models vary in size: GPT-4 comprises approximately 1.7 trillion parameters, PaLM 2 has 340 billion, and MPT Instruct is the smallest with 7 billion parameters. This diverse range enables us to evaluate the influence of model size on performance. For each model, we conducted experiments using two temperature settings (0.2 and 0.7) to examine the effects of randomness on model performance. Every model was run twice at each temperature setting. However, users cannot set a random seed for GPT-4 and PaLM 2, resulting in varying outputs between runs. In contrast, MPT Instruct was executed twice using two distinct random seeds.

\subsection*{Prompts}
The quality of prompts used to guide LLMs significantly impacts their performance, akin to the instructions given to crowdworkers. We tailored the instruction around the prompt set up to focus on understanding the financial context around each question. Each input prompt comprises: (1) textual description of the task, (2) a sentence with highlighted entities\footnote{We indicate the locations of both entities of interest by adding ** before and after entity1 and \_\_ before and after entity2.}, and (3) a numbered list of relation options (labels) specific to the entity pair. To avoid bias towards particular label orderings,
 we shuffle the option list. 
We experimented with 6 distinct prompt types which fall into 3 categories: zero-shot, few-shot and few-shot chain-of-thought (CoT) prompts. 
These prompts are based on the annotation instructions provided to MTurk\footnote{\url{https://docs.aws.amazon.com/pdfs/AWSMechTurk/latest/AWSMechanicalTurkRequester/amt-dg.pdf}} crowdworkers for the REFinD annotations (taken from \citet{10.1145/3477495.3532019}), facilitating a better comparison with their outputs. 

For zero-shot prompts, we used: (1) \textit{simple prompt}, a brief task description in basic English and (2) \textit{full instruction prompt}, an extended version with a more comprehensive task description from the REFinD MTurk annotation instructions, an example of this is provided in Figure \ref{fig:example TDP}. Few-shot prompts, include: (3) \textit{1-shot} and (4) \textit{5-shot}, which build upon the full instruction prompt by adding a few task examples, tailored to the specific entity-pair type.  Lastly, we experimented with few-shot CoT prompts: (5) \textit{1-shot CoT} and (6) \textit{5-shot CoT}. CoT prompts incorporates both the task descriptions and examples, as well as the reasoning behind each example's decision, as this approach has proven beneficial for other annotation tasks~\cite{wei2022chain}. 


\begin{figure}[h]
\footnotesize
 \begin{framed}
\textbf{}Select date of formation relationship described in one sentence. Given a single sentence: The predecessor **Mississippi Power Company** was incorporated under the laws of the State of Maine on November 24, 1924 and was admitted to do business in Mississippi on \_\_December 23, 1924\_\_ and in Alabama on December 7, 1962. With 2 highlighted phrases: Mississippi Power Company and December 23, 1924. Select a multiple choice answer from options below, which best describes the relation between Mississippi Power Company and December 23, 1924. 
\\
\\Please choose the MOST appropriate relation from the following options:
\begin{enumerate}
    \item Mississippi Power Company is/was formed on December 23, 1924
    \item Mississippi Power Company is/was acquired on December 23, 1924
    \item no/other relation between Mississippi Power Company and December 23, 1924
\end{enumerate}
 \end{framed}
 \caption{Full instruction prompt example. } 
 \label{fig:example TDP}
\end{figure}

\subsection{Evaluation} \label{evaluation}
We assess the performance in comparison to expert annotators using accuracy and micro-averaged F1 scores. These metrics are calculated separately for each entity pair, and we report the mean average across entity pairs. Since each model's experiment is run twice, we also average these metrics from the two runs and report this as the final metric. Additionally, we measure the agreement between experiments, the time and cost of annotations, and the reliability index to analyze the efficiency and robustness of LLMs as annotators.


   
\subsubsection{Inter Annotator Agreement (IAA)}
\mbox{}\\
We evaluate the agreement between different experiment settings to capture the model's self-consistency and assess the quality and reliability of the annotations. This metric demonstrates how uniformly annotators interpret the given task.   
To calculate the agreement between two annotators, we use Cohen's Kappa~\cite{cohen1960coefficient} and for agreement among more than two annotators, we use Fleiss' Kappa~\cite{fleiss1971measuring}.

\begin{table*}[t!]
\centering
\resizebox{\textwidth}{!}{
\begin{tabular}{clccccccc}
\hline
\multicolumn{9}{c}{\textbf{Micro-Averaged F1 Score/ Accuracy(\%)}} \\ \hline
\multicolumn{1}{c|}{} & \multicolumn{1}{c|}{} & \multicolumn{1}{c|}{} & \multicolumn{2}{c|}{\textbf{Zero-Shot Prompt}} & \multicolumn{2}{c|}{\textbf{Few-Shot Prompt}} & \multicolumn{2}{c}{\textbf{Few-Shot CoT Prompt}} \\ \cline{4-9} 
\multicolumn{1}{c|}{\multirow{-2}{*}{\textbf{Annotator}}} & \multicolumn{1}{c|}{\multirow{-2}{*}{\textbf{Type}}} & \multicolumn{1}{c|}{\multirow{-2}{*}{\textbf{\begin{tabular}[c]{@{}c@{}}Temperature \\ Setting\end{tabular}}}} & \multicolumn{1}{c|}{\textbf{simple prompt}} & \multicolumn{1}{c|}{\textbf{full instruction}} & \multicolumn{1}{c|}{\textbf{1-shot}} & \multicolumn{1}{c|}{\textbf{5-shot}} & \multicolumn{1}{c|}{\textbf{1-shot CoT}} & \textbf{5-shot CoT} \\ \hline
 & GPT-4 & \multicolumn{1}{c|}{\cellcolor[HTML]{FFFFFF}0.2} & \cellcolor[HTML]{FFFFFF}67.4/63.4 & 68.5/64.6 & 65.0/60.1 & 67.6/63.8 & 64.5/58.4 & 68.4/\textbf{65.4} \\
 & GPT-4 & \multicolumn{1}{c|}{0.7} & \textbf{67.6/63.6} & 68.4/64.6 & 65.0/60.0 & 67.7/63.9 & 64.6/\textbf{58.4} & 68.4/\textbf{65.4} \\
 \cline{2-9}
 & PaLM 2 & \multicolumn{1}{c|}{0.2} & 62.3/53.9 & 62.2/53.8 & 66.4/60.1 & 66.0/59.2 & 64.7/55.9 & 65.6/57.2 \\
 & PaLM 2 & \multicolumn{1}{c|}{0.7} & 64.5/56.0 & 64.4/56.0 & \textbf{67.3/60.9} & \textbf{68.7}/63.8 & 64.9/57.4 & 65.9/59.2 \\
 \cline{2-9}
 & MPT Instruct & \multicolumn{1}{c|}{0.2} & 20.0/21.9 & 31.1/27.6 & 18.6/18.0 & 42.5/36.7 & 20.1/18.5 & 45.2/36.1 \\
\multicolumn{1}{c}{\multirow{-3}{*}{LLM}}  & MPT Instruct & \multicolumn{1}{c|}{0.7} & 20.8/24.7 & 24.8/27.3 & 22.7/24.2 & 30.5/31.1 & 22.2/23.2 & 33.9/30.8 \\ \cline{2-9} \noalign{\vskip\doublerulesep\vskip-\arrayrulewidth} \cline{2-9}
 & Ensemble (All LLMs) & \multicolumn{1}{c|}{0.2} & 65.2/60.1 & 66.0/60.7 & 63.9/58.1 & 68.1/63.3 & 63.3/56.4 & \textbf{68.8}/63.8 \\ 
 & Ensemble (GPT-4 w Palm 2) & \multicolumn{1}{c|}{0.2} & 67.2/63.2 & \textbf{68.6/64.7} & 65.0/60.1 & 67.8/\textbf{64.0} & 64.3/58.1 & 68.2/65.2 \\ 
 & Ensemble (GPT-4 w MPT Instruct) & \multicolumn{1}{c|}{0.2} & 67.2/63.2 & \textbf{68.6/64.7} & 65.0/60.1 & 67.8/\textbf{64.0} & 64.3/58.1 & 68.2/65.2 \\ 
 & Ensemble (Palm 2 w MPT Instruct) & \multicolumn{1}{c|}{0.2} & 62.6/54.3 & 61.9/53.6 & 66.7/60.5 & 66.1/59.4 & 64.5/55.7 & 65.4/56.9 \\ \hline
Human & \multicolumn{1}{l}{Mturk Annotators} & \multicolumn{1}{c|}{-} & - & 38.6/40.7 & - & - & - & - \\ \hline
\end{tabular}
}
\caption{ Annotator performance in terms of micro-averaged F1-Score and accuracy against expert assigned labels.}
\label{llm_table1}
\end{table*}

\subsubsection*{Reliability Index (LLM-RelIndex)}
\mbox{}\\
To aggregate the label for each sample from multiple annotators, we could simply calculate the raw voting counts for each label from K annotators.  However, this  approach has an issue when annotators all choose distinct labels, then an arbitrary label would be selected. As those distinct labels could be semantically related, such as \textsc{member of}, \textsc{employee of} and \textsc{founder of}, incorporating such label similarity can improve the aggregation precision. Thus, we refine the voting approach by taking label similarity into account i.e., the similarities between its assessments $a_i$ and each label $l$. 
The refined voting score, which considers the assessments of multiple annotators, measures the agreement for each label $l$ as $\text{vote}(i,l) = sim(a_i, l)$. We then define the confidence as $\text{confid}(l) = \frac{1}{K} \sum_{i=1}^{K} \text{vote}(i,l)$. Note that similarity is defined as per the judgements of domain experts. 

Additionally, we introduce the Reliability-Index, defined as the maximum confidence score $\text{confid}(l)$ of the label $l$:
\begin{equation}\label{eq:3}
\text{LLM-RelIndex}_i = \arg\max_{l \in L} \text{confid}(l)
\end{equation}

The Reliability-Index aids in identifying the most reliable label for each instance.
It enables the detection of outputs that warrant human expert attention. 

\subsubsection*{Time \& cost}\label{sec:time}

\mbox{}\\
For models served via API, the price per instance depends on the number of tokens (GPT-4\footnote{\url{https://openai.com/pricing}, Accessed on 31/07/2023, GPT-4 8K context input price: \$0.03/1K tokens, output price: \$0.06/1K tokens. Number of tokens was calculated using the tiktoken package.}) or characters (PaLM 2\footnote{\url{https://cloud.google.com/vertex-ai/pricing}, Accessed on 31/07/2023, PaLM 2 Text Bison: \$0.0010/1K characters.}) in both the prompt and generated outputs. Consequently, the annotation cost was calculated by multiplying the average number of tokens/characters in the prompt and output, the number of instances, and the price per instance.
For the open-source MPT-Instruct model, the cost was based on the per-hour price of the AWS machine utilized. Due to high GPU memory requirements, we used \textit{p3.2xlarge} machines with 1 Tesla V100 GPU\footnote{https://aws.amazon.com/ec2/instance-types/p3/}. The annotation cost was calculated by multiplying the average time taken per instance in hours, the number of instances, and the price per hour.

\section{Results}
In this section, we discuss our experimental findings, focusing on model performance, annotator agreement, error analysis and reliability.

\subsection{Model Performance}
\begin{figure}[h]
    \centering
    \includegraphics[width=0.45\textwidth]{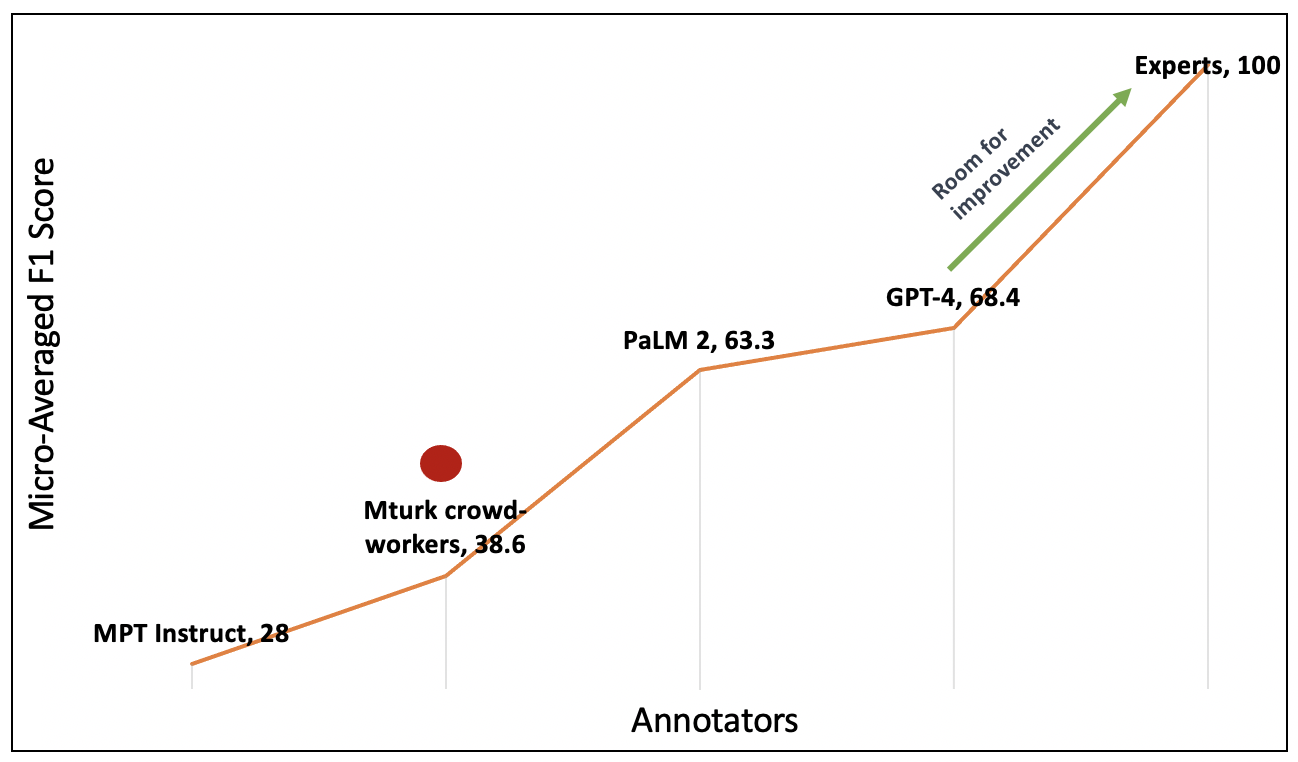}
    \caption{Annotator performance in terms of micro-averaged F1-Score under \textit{full instruction prompt}.}
    \label{fig:enter-label}
\end{figure}

Table \ref{llm_table1} presents the micro-averaged F1 score and accuracy for each LLM by prompt type and temperature setting, as well as the performance of MTurk annotators.
We observe that GPT-4 and PaLM 2 significantly outperform crowdsourced annotations, with a margin of up to 29\%. Both models exhibit comparable performance, with GPT-4 being the best. MPT Instruct demonstrates lower overall performance but still outperforms the human annotators in terms of F1-score when using \textit{5-shot CoT prompt}. These results highlight the potential of LLMs as annotators. However, none of the models reach the expert performance, indicating that domain-specific settings still require expert's involvement. 
Figure \ref{fig:enter-label} visualizes the results for the \textit{full instruction prompt}, which is identical to the MTurk instructions. 

Regarding the impact of prompt type on model performance, Table \ref{llm_table1} reveals that the input prompt design significantly influences LLM performance. GPT-4 and PaLM 2 exhibit higher robustness under different prompts (5-7\% difference), whereas prompt type has a strong effect on MPT Instruct performance (19\%). MPT Instruct benefits considerably from additional examples (\textit{5-shot} and \textit{5-shot CoT}). Interestingly, few-shot and few-shot CoT prompts do not consistently outperform the zero-shot \textit{full instruction prompt}. GPT-4 achieve its highest micro-averaged F1 score using the zero-shot \textit{full instruction prompt}. 

Comparing performance at 0.2 and 0.7 temperature settings, we find that GPT-4 and PaLM 2 outputs remains stable regardless of the randomness introduced by the temperature parameter. While PaLM 2 consistently exhibits higher performance at 0.7, the observed performance differences are not statistically significant at the 0.05 significance level using a two-tailed t-test \ref{tab:statistical_test}. 
MPT Instruct performance is heavily affected by temperature settings, but no consistent pattern of superiority emerges for either setting. The highest scores are achieved at 0.2 with 5-shot example prompts. 

Additionally, we evaluate the performance of an ensemble of models using a simple majority voting approach, which mimics having multiple annotators. While this approach results in the highest overall accuracy score, it does not consistently improve performance across all prompt types compared to a single model approach.

\begin{table*}[t!]
\centering
\resizebox{0.7\textwidth}{!}{%
\begin{tabular}{r|cc|cc|cc}
\hline
\multirow{2}{*}{LLM} &
  \multicolumn{2}{c|}{Zero-Shot Prompt} &
  \multicolumn{2}{l|}{Few-Shot Prompt} &
  \multicolumn{2}{l}{Few-Shot CoT Prompt} \\ \cline{2-7} 
 &
  \multicolumn{1}{l}{simple} &
  \multicolumn{1}{l|}{full instruction} &
  \multicolumn{1}{l}{1-shot} &
  \multicolumn{1}{l|}{5-shot} &
  \multicolumn{1}{l}{1-shot CoT} &
  \multicolumn{1}{l}{5-shot CoT} \\ \hline
GPT-4        & 69.3 & 70.2 & 71   & 68.8 & 72.5 & 66.2 \\
PaLM 2        & 74.5 & 73.8 & 74.9 & 76.1 & 79.8 & 80.7 \\
MPT Instruct & 46.4 & 52.5 & 48.4 & 57.5 & 49.7 & 64.9
\end{tabular}
}
\caption{Proportion of LLM Hallucinations for instances labelled as \textsc{no/other relation} by experts} 
\label{tab:llm_hallucination_prop}
\end{table*}

\subsection{Inter-Annotator Agreement}
High performance alone is insufficient for LLMs to serve as annotators, their output must also be consistent to be considered reliable. Therefore, we assess the consistency of the output by measuring agreement scores for models in different experiment settings shown in Table \ref{tab:iaa1}.  
First, we evaluate whether the models produce consistent outputs with the exact same parameters. For each experiment setting, we measure the IAA between the two runs of each model and then present an average score (row 1).  

We observe that none of the models exactly replicate the outputs. GPT-4 and PaLM 2 exhibit high levels of agreement, while MPT runs with two different random seeds display significant differences. 

\begin{table}[h]
\centering
\resizebox{0.9\columnwidth}{!}{%
\begin{tabular}{lccc}
\hline
 & \multicolumn{1}{l}{GPT-4} & \multicolumn{1}{l}{PaLM 2} & \multicolumn{1}{l}{MPT} \\ \hline
Random seed run1 vs run2 & \textbf{0.95} & 0.88 & 0.395 \\
Temperature 0.2 vs 0.7 & \textbf{0.95} & 0.85 & 0.30 \\
\hline
Zero-shot: simple vs full & \cellcolor[HTML]{FFFFFF}0.87 & \cellcolor[HTML]{FFFFFF}\textbf{0.88} & \cellcolor[HTML]{FFFFFF}0.39 \\
Few-shot: 1- vs 5-shot & \cellcolor[HTML]{FFFFFF}\textbf{0.84} & \cellcolor[HTML]{FFFFFF}0.79 & \cellcolor[HTML]{FFFFFF}0.28 \\
Few-shot CoT: 1- vs 5-shot & \cellcolor[HTML]{FFFFFF}0.8 & \cellcolor[HTML]{FFFFFF}\textbf{0.82} & \cellcolor[HTML]{FFFFFF}0.28 \\
\hline
All prompts (Fleiss) & \cellcolor[HTML]{FFFFFF}\textbf{0.83} & \cellcolor[HTML]{FFFFFF}0.79 & \cellcolor[HTML]{FFFFFF}0.31\\
\hline
\end{tabular}%
}
\caption{Pairwise IAA in terms of Cohen Kappa (top 5 rows) and IAA between outputs for all prompts in terms of Fleiss Kappa (last row). First two rows present mean averaged values of pairwise Cohen Kappa for each prompt type.}
\label{tab:iaa1}
\end{table}

\noindent We then evaluate the agreement between outputs produced under two different temperature settings (row 2). GPT-4 agreement remains high even when varying the temperature parameter, while scores of PaLM 2 and MPT decrease. 
Furthermore, we compare the agreement between outputs produced using different prompts, both pairwise (using Cohen's Kappa, rows 3-5) and between the group of prompts (using Fleiss Kappa, row 6). We find that the choice of prompt has a more substantial impact on the outputs of the model, reducing the agreement for all LLMs. 
Overall, GPT-4 and PaLM 2 demonstrate reasonably high agreement across various experiment settings, indicating their overall reliability for the annotation task.

\subsection{Error Analysis}

In our error analysis, we aim to identify and categorize common issues encountered by LLMs during the annotation process. By examining instances with incorrect answers, hallucinated relations, and confident misannotations, we aim to gain insights into the challenges faced by LLMs and explore potential improvements for their performance in complex tasks, such as relation extraction.\\
\subsubsection{Semantic Ambiguity}
\mbox{}\\
We analyze instances where LLMs return incorrect answers and observe that these errors often stem from the proximity and similarity of the answer options, causing confusion in identifying the most accurate response. Common trends include \textsc{member of} instead of \textsc{employee of} and \textsc{formed in} rather than \textsc{operations in}. This highlights the need to improve LLM's comprehension of subtle differences. For instance, in the example ``\underline{\textcolor{red}{W. Howard Keenan , Jr.}} has served as a director of \underline{\textcolor{blue}{Midstream Management}} since February 2014", both GPT-4 and PaLM 2 incorrectly choose \textsc{member of} over the correct relation \textsc{employee of}. Although MPT Instruct's result is also inaccurate, its answer varies significantly by prompt type, exhibiting a level of randomness not observed in the other two LLMs. Its also worth noting that MPT Instruct returns blanks for some instances.   0.5\% of the responses from MPT Instruct for each prompt variation were blanks.
\subsubsection{Relation Hallucinations}
\mbox{}\\
In our relation extraction task, we provide the LLMs with limited label options, including an option for \textsc{no/other relation} available for every entity pair.  Consequently, we expect minimal instances of hallucinations, i.e., LLMs inventing new relations between specified entities not present in the label set or generating off-topic responses. We analyze the LLM outputs for instances labeled as  \textsc{no/other relation} by the experts and report the proportion of hallucinations among them (Table \ref{tab:llm_hallucination_prop}). We observe that  hallucinations primarily emerge from PaLM 2 for \textit{5-shot CoT}, where 80.7\% of instances labeled as \textsc{no/other relation} by the experts were misidentified by PaLM 2 as hallucinations. Overall, LLMs exhibit a higher tendency to generate new relations when the expert label is \textsc{no/other relation}. GPT-4 and PaLM 2 tend to hallucinate more than MPT Instruct. We post-process the hallucinated relations to extract relation styles similar to those in the label options. The most common relations extracted from these are \textsc{agreement with}, \textsc{shares of}, \textsc{member of} and \textsc{subsidiary of}. 
\begin{figure*}[t!]
\footnotesize
 \begin{framed}
 \vspace{-0.3cm}
\textbf{Scenario 1 (Crowdworkers incorrect, LLMs correct):\\}
\textbf{Instance}: \underline{\textcolor{red}{Personal Lines}} underwriting profit for the three months ended September 30 , 2017 was \underline{\textcolor{blue}{\$ 40.8 million}}, compared to \$ 23.3 million for the three months ended September 30 , 2016 , an improvement of \$ 17.5 million.\\
\vspace{-0.3cm}
\\
\textbf{Expert Label:} \textcolor{purple}{\textsc{Profit of}} \\
\textbf{Crowdworker Label:} \textsc{Profit of, No/Other Relation , Loss of}\\
\textbf{LLMs Label:} \textcolor{purple}{\textsc{Profit of}} \\
\vspace{-0.3cm}
\hrule
\vspace{0.1cm}
\textbf{Scenario 2 (Crowdworkers and LLMs incorrect):\\} 
\textbf{Instance}: Our \underline{\textcolor{red}{Hawaii Gas}} entered into licensing agreements with Utility Service Partners , Inc. and America's Water Heater Rentals , LLC , both indirect subsidiaries of \underline{\textcolor{blue}{Macquarie Group Limited }}, to enable these entities to offer products and services to Hawaii Gas's customer base. \\
\vspace{-0.3cm}
\\
\textbf{Expert Label:} \textcolor{purple}{\textsc{Subsidiary of}} \\
\textbf{Crowdworker Label:} \textsc{No/Other Relation, Subsidiary of, Shares of}  \\
\textbf{LLMs Label:} \textsc{Agreement with} \\
\vspace{-0.3cm}
\hrule
\vspace{0.1cm}
\textbf{Scenario 3 (Crowdworkers correct, LLMs incorrect):\\} 
\textbf{Instance}: On December 10 , 2014 , Orbital Tracking Corp. purchased certain contracts from Global Telesat Corp , a Virginia corporation ( GTC ) for \$ 250,000 pursuant to an asset purchase agreement by and among \underline{\textcolor{red}{Orbital Tracking Corp.}}, its wholly owned subsidiary Orbital Satcom, GTC and World Surveillance Group , Inc. ( World ) , \underline{\textcolor{blue}{GTC}}'s parent.\\
\vspace{-0.3cm}
\\
\textbf{Expert Label:} \textcolor{purple}{\textsc{Subsidiary of}} \\
\textbf{Crowdworker Label:} \textcolor{purple}{\textsc{Subsidiary of}}\\
\textbf{LLMs Label:} \textsc{Agreement with}
\vspace{-0.2cm}
 \end{framed}
 \caption{Error Analysis: Qualitative examples illustrating different scenarios of how MTurk Crowdworkers and LLMs demonstrated high confidence on incorrect answer choices. }
 \label{fig:scenario}
\end{figure*}

\subsubsection{Confident Misannotations}
\mbox{}\\
We analyze instances where LLMs and crowdworkers return incorrect answers with high confidence (answers selected by majority of annotators). The relationship between high confidence and incorrect answer choice varies, and we observe three scenarios: (i) the majority of crowdworker labels are incorrect while the majority of LLM labels are correct, (ii) the majority of both crowdworker and LLM annotations are incorrect, and (iii) the majority of crowdworker labels are correct while the majority of LLM labels are incorrect. Qualitative examples of these can be found in Figure \ref{fig:scenario}.
This analysis demonstrates that the varying dynamics between LLMs and crowdworkers emphasize the importance of refining LLMs to better understand nuanced distinctions and improve their reliability in annotation tasks. Furthermore, the analysis highlights the potential benefits of combining the expertise of both LLMs and human annotators to achieve more accurate and reliable annotations in complex tasks, such as relation extraction.

\subsection{LLM-RelIndex Based Accuracy Analysis}

\begin{figure}
   \centering
   \includegraphics[width=\columnwidth]{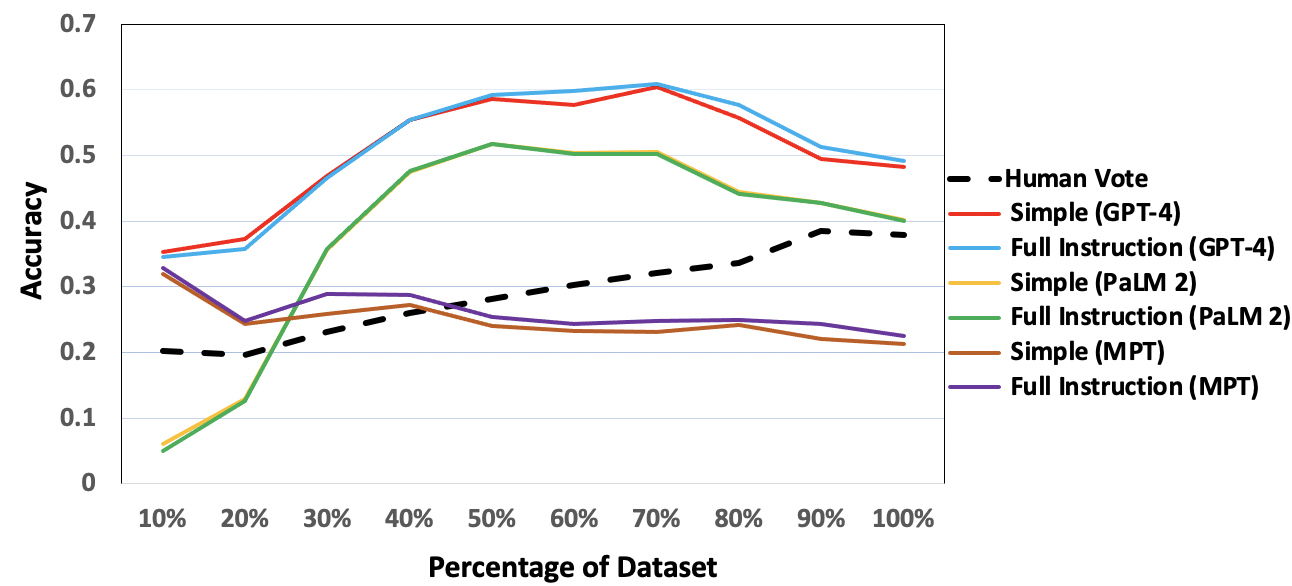}
   \caption{Human vs LLMs at Zero-shot using \emph{LLM-RelIndex}}
   \label{fig:LLM-RelIndex1}
\end{figure}

\begin{figure}
   \centering
   \includegraphics[width=\columnwidth]{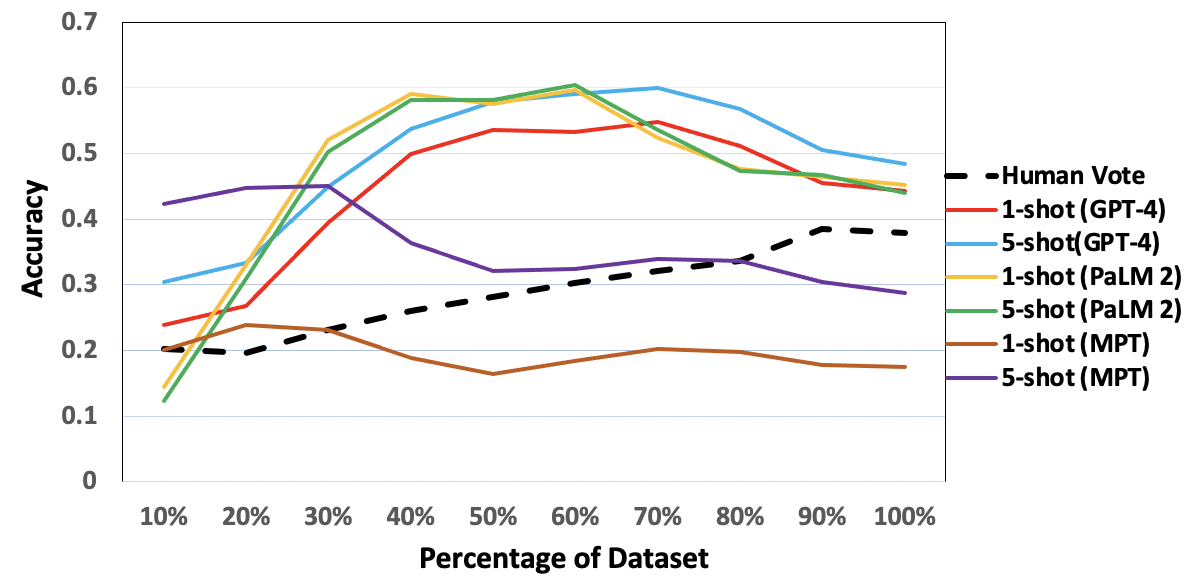}
   \caption{Human vs LLMs at Few-shot using \emph{LLM-RelIndex}}
   \label{fig:LLM-RelIndex2}
\end{figure}

\begin{figure}
   \centering
   \includegraphics[width=\columnwidth]{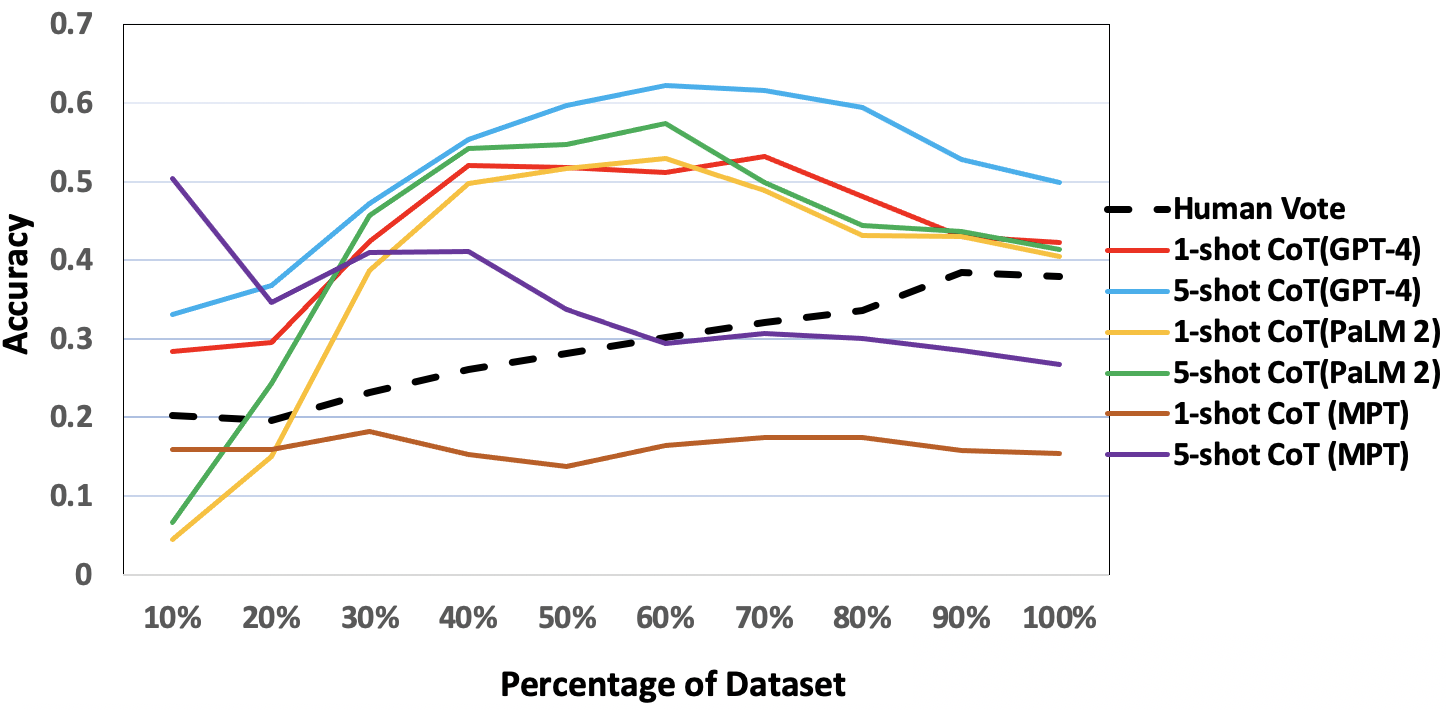}
   \caption{Human vs LLMs at Few-shot CoT using \emph{LLM-RelIndex}}
   \label{fig:LLM-RelIndex3}
\end{figure}

In this analysis, we employ the LLM-RelIndex majority voting scheme to assess the accuracy derived from human votes and LLM results across all six prompt variations on the dataset. The data is arranged in descending order of LLM-RelIndex that is we moved from instances that were simple  to annotate to the more complex ones and we present the accuracy for incremental percentages of the dataset. We showcase the plots for three distinct cases: (i) zero-shot (Figure \ref{fig:LLM-RelIndex1}), (ii) few-shot (Figure \ref{fig:LLM-RelIndex2}), and (iii) few-shot CoT (Figure \ref{fig:LLM-RelIndex3}).

Our observations indicate that for all three cases, GPT-4 and PaLM 2 outperform both Human Votes and MPT Instruct when considering $\sim$65\% of the dataset. However we also observed a drop in accuracy in the top 20\% of the dataset where there were high level agreements among LLMs. This can be attributed to the instances which were simple to annotate but easier to error on. Hence we observe that in those instances most of the LLMs made the same mistakes as human annotators which were inconsistent with expert choices. For example ``The number of shares that are sold by Cowen after delivering a sales notice will fluctuate based on the market price of \underline{\textcolor{red}{Dermira, Inc}} common stock during the sales period and limits Dermira, Inc. set with \underline{\textcolor{blue}{Cowen}}." Most of the LLMs chose \textsc{Agreement with} over \textsc{Shares of} where the latter is the correct relation.

Additionally, PaLM 2's performance exhibits an upward trend, as we transition from zero-shot to few-shot, and ultimately to few-shot CoT scenarios. We also find that all LLMs demonstrate improved results for \textit{5-shot} and \textit{5-shot CoT}, suggesting that having more examples and explanations enhances the reliability of LLM-generated annotations.

As we progress towards complete dataset coverage, again we see a decline in performance is noted. This outcome is anticipated since instances with lower LLM-RelIndex scores become more prevalent as we approach more complex instances. Here the LLMs likely lack  confidence in relations between specific entity pairs.

Overall, LLM-RelIndex allows us to confidently assert that LLMs can serve as more reliable annotators for $\sim$65\% of this dataset. For cases beyond this threshold, expert intervention is necessary to determine the appropriate annotation. This strategy effectively reduces the cost and time associated with human annotation of the entire dataset, streamlining the process considerably.

\subsection{Time and Cost Analysis} 
\label{sec:time}
We calculate the annotation cost for each of the LLMs (detailed in Evaluation section) and compare it to our estimated cost of MTurk annotations. The average input prompt size ranges from 191 tokens (814 characters) for \textit{simple prompts} to 441 tokens (1954 characters) for \textit{5-shot CoT prompts}. On average, GPT-4 generates an output of 17 tokens (65 characters) for all prompt types. The outputs of PaLM 2 vary more in size, from 70 to 36 tokens (298 and 147 characters), with shorter outputs for longer prompts like few-shot and few-shot CoT. Each model can process an instance within 1-5 seconds, with longer prompts requiring more processing time. MPT Inference on average takes 0.96 seconds for \textit{simple prompts} and 1.81 for the longest \textit{5-shot CoT prompts}. The annotation price increases with the prompt size, and for our dataset of 3598 instances, it ranges from \$24-51 for GPT-4, \$5-9 for PaLM 2  and \$29-55 for MPT Instruct. 

For crowdsourced human annotators, the time and associated cost would be higher. Assuming a human annotator takes 45 seconds per instance and is paid the US minimum wage of \$7.25 per hour\footnote{\url{https://www.dol.gov/agencies/whd/minimum-wage}, Accessed on 31/07/2023.}, the dataset's annotation cost using a single annotator amounts to \$389. However, the crowdsourced annotation process typically involves multiple annotators per instance.  
These outcomes demonstrate that automated annotations are more efficient in terms of time and cost compared to human labelling.

\section{Discussion}
In this section we discuss our findings and share recommendations for future annotation tasks. 
Our experiments demonstrated the potential of LLMs as data annotators for tasks within the financial domain. Specifically, GPT-4 and PaLM 2 have exhibited exceptional performance, surpassing the accuracy of the non-expert crowdworkers, while delivering time and cost savings. PaLM 2 has achieved comparable results to GPT-4, despite its smaller size, at a fraction of the cost ($\sim$5 times less).  These models have also displayed robustness by producing consistent outputs across various parameter and prompt configurations. However, it is crucial to recognize that LLMs' performance does not yet match that of domain experts and expert involvement remains necessary for obtaining high-quality annotations with minimal or no noise.

The next generation of annotation approaches in domain-specific contexts should consider adopting a hybrid strategy, harnessing both automated and expert-generated annotations to optimize results. In these settings, approximating model uncertainty, e.g., via the LLM-RelIndex, can help prioritize instances that require expert attention. In all annotation tasks, the ability to formulate detailed instructions is a vital factor, regardless of whether annotators are human or LLMs. Carefully crafting prompts, guided by an understanding of the task and the specific LLM being used helps optimize the outputs generated by the LLMs. 
 

We, therefore, recommend that researchers conduct small preliminary experiment on a data subset to assess model capabilities and identify optimal parameter and prompt configurations. The specifics of the task should inform researchers about the \textit{tolerance for annotation noise}, 
 allowing them to train new models using automatically annotated data accordingly.  
Moreover, future annotation tasks can benefit from more open task formulations, leveraging the generative abilities of LLMs. 
For instance, in our task, LLMs have the potential to help identify more relations than the original pre-defined set. As such, future experiment can be done to check if these LLM-annotated data boost downstream performances. Lastly, it is essential to remain mindful that model biases may differ from those of crowdworkers and to account for these differences where necessary. 
\section{Limitations}
One of the main limitations of this work is that the evaluation is performed only on a single dataset, covering a single task. The dataset contains the texts from one particular source, SEC filings, and it would be interesting to compare the results when the texts come from other financial sources, such as news or earning calls. This limitation partially comes from the costs of using the LLMs, and partially from the absence of financial datasets with annotations produced by individual crowdworkers released publicly. 

In this work we present the breakdown of the results and their analysis by relation categories in the Appendix due to the page limit. We found that model performance varies strongly between the entity pair groups similar to \citet{10.1145/3477495.3532019} with \textsc{organization-organization} being the most challenging category. In future work, we aim to expand our analysis further with respect to categories of errors frequently associated with this task and financial domain such as numerical inference, semantic and directional ambiguity.

We observe that our LLM-Rellndex metric is subject to error, particularly with instances that are easy to annotate. Efforts are underway to enhance this metric. Furthermore, we are exploring the adoption of an automated and systematic approach for calculating similarity scores rather than depending on experts' judgment. Additionally, we intend to incorporate multi-label samples into our approach, given that some similar labels may closely align for some cases.

Finally, while providing the discussion, we do not experimentally demonstrate how the automatically annotated dataset can be used, either to improve relation extraction model performance, or to develop smaller efficient models. We recognize the importance of this and leave this to future work.

\section{Conclusion}
In this study, we have showcased the remarkable potential of using LLMs as a robust alternative to non-expert crowdworkers for domain-specific task by comparing three LLMs of varying sizes. Due to large volume of unstructured documents within financial domain, leveraging LLMs for annotations significantly reduces the time spent by humans on manual annotation, while providing valuable insights for making well-informed downstream decisions and driving efficient business outcomes.  Our evaluation shows that larger models like GPT-4 and PaLM 2 excel in these tasks, while incorporating more examples into prompts for smaller models like MPT Instruct can yield improved results. We also introduced the reliability index, a metric that identifies reliable labels and detects outputs requiring expert attention, enhancing quality control and decision-making. Our error analysis provides valuable insights for future improvements.

The integration of LLMs streamlines the annotation process, delivering consistent, high-quality outputs that result in substantial time savings and cost-effectiveness. However, their performance does not yet match that of experts who possess a nuanced understanding of the subject matter. While LLMs offer scalability and reduced time and costs compared to employing experts, there exists a trade-off between the convenience and efficiency of LLMs and the precision provided by expert annotators. Consequently, the decision to employ LLMs as annotators should be carefully guided by the desired level of accuracy and the complexity of the task at hand, striking the right balance between automation and human expertise.

\section{Acknowledgments}
We would like to thank Armineh Nourbakhsh, Natraj Raman, Xiaomo Liu, Manuela Veloso, and our anonymous reviewers for their thoughtful comments and feedback which greatly contributed to the quality of this work.

Disclaimer. This paper was prepared for informational purposes by the Artificial Intelligence Research group of JPMorgan Chase \& Co. and its affiliates (“JP Morgan”), and is not a product of the Research Department of JP Morgan. JP Morgan makes no representation and warranty whatsoever and disclaims all liability, for the completeness, accuracy or reliability of the information contained herein. This document is not intended as investment research or investment advice, or a recommendation, offer or solicitation for the purchase or sale of any security, financial instrument, financial product or service, or to be used in any way for evaluating the merits of participating in any transaction, and shall not constitute a solicitation under any jurisdiction or to any person, if such solicitation under such jurisdiction or to such person would be unlawful.

\section{Ethical Considerations}
 
This paper explores the use of LLMs for data annotation. As such, the prevailing concerns around the use of LLMs apply to this work.  This includes the potential to generate text containing bias, stereotypes, misinformation and, as noted in the discussion, hallucinations.  Outside of issues concerning LLM usage, we do not anticipate other ethical concerns with this work.

\nocite{*}
\section{Bibliographical References}\label{sec:reference}

\bibliography{aaai24}

\appendix

\label{sec:appendix}
\section{Appendices}
\hspace{0.2cm}

\subsection{Dataset relation distribution}
\hspace{0.5cm}
\begin{table}[h!]
\centering
\scalebox{1.1}{
\begin{tabular}{cc}
\hline
Entity-Pair & \begin{tabular}[c]{@{}c@{}}No. of \\ Instances\end{tabular} \\ \hline
ORG-GPE & 710 \\
ORG-ORG & 913 \\
ORG-DATE & 554 \\
ORG-MONEY & 281 \\
PER-ORG & 485 \\
PER-TITLE & 655 \\ \hline
Total & 3598 \\ \hline
\end{tabular}
}
\caption{Dataset Relation Distribution}
\label{tab:relatn_dist}
\end{table}

\vspace{32mm}
\subsection{Metrics for MTurk Annotators}
\vspace{5mm}
\begin{table}[h!]
\centering
\begin{tabular}{|lc|}
\hline
\multicolumn{2}{|c|}{\textbf{Micro F1 Score/ Accuracy (\%)}}           \\ \hline
\multicolumn{1}{|c|}{\textbf{Entity Pair}} & \textbf{MTurk Annotators} \\ \hline
ORG-GPE                                    & 37.3/35.8                 \\
ORG-ORG                                    & 13.5/21.6                 \\
ORG-DATE                                   & 31.4/45.0                 \\
ORG-MONEY                                  & 26.4/29.1                 \\
PER-ORG                                    & 33.9/32.3                 \\
PER-TITLE                                  & 89.0/80.4                 \\
Total                                      & 38.6/40.7                 \\ \hline
\end{tabular}
\caption{MTurk Annotator Micro-average F1 Score/Accuracy by Entity Pair}
\end{table}


\subsection{LLM Setup and Configuration}


The setup and configuration of each LLMs have some overlap such as specifying the location of each entity in the text, however, there are notable differences as well. These differences enabled each LLM to perform at its best. Figure 1 explains the different piece of LLM setup and configuration. Unlike GPT-4, where we had "system role", in PaLM 2 we had "Additional Instruction". This is the unique prompt design for the different prompt type.

\begin{figure}[h!]
\footnotesize
 \begin{framed}
\textbf{Setup \& Configuration: ORG-DATE} \\
\underline{\textbf{\textsc{Instruction}}}:Select the statement that best describes the relation in the example sentence below. Ignore any grammatical errors. If there are multiple options, please choose the one that is clearest and most obvious from the sentence. \\
\\
\underline{\textbf{\textsc{Prompt}}}: The predecessor \underline{\textcolor{red}{Mississippi Power Company}} was incorporated under the laws of the State of Maine on November 24, 1924 and was admitted to do business in Mississippi on  \underline{\textcolor{blue}{December 23, 1924}} and in Alabama on December 7, 1962. \\
\\
\underline{\textbf{\textsc{Prompt+}}}: Please choose the MOST appropriate relation from the following options:
\begin{enumerate}
    \item \textcolor{red}{Entity1} is/was formed on \textcolor{blue}{Entity2}.
    \item \textcolor{red}{Entity1} is/was acquired on \textcolor{blue}{Entity2}.
    \item No/other relation between \textcolor{red}{Entity1} and \textcolor{blue}{Entity2}.
\end{enumerate} 
\underline{\textbf{\textsc{System role}}}: You are an AI assistant and relation extraction checker. You read the prompt, note where the entities in question are and determine the relation between them. Once done, please select from option which best suite the relation.
 \end{framed}
 \end{figure}
\vspace{27mm}

\begin{figure*}[h!]
\footnotesize
 \begin{framed}
\textbf{GPT-4 setup follows:} 
Using the context from "setup piece and configuration" \\
\textbf{Zero-shot:} \\
This starts with \textsc{\textbf{Prompt}}, followed by \textsc{\textbf{Prompt+}} and finally \textsc{\textbf{System role}} and \textsc{\textbf{Response}}.\\
\textbf{Few-shot:} \\
This starts with \textsc{\textbf{Instruction}}, followed by \textsc{\textbf{Prompt with example(s)}}, \textsc{\textbf{Prompt+}} and finally \textsc{\textbf{System role}} and \textsc{\textbf{Response}}. \\
\textbf{Few-shot CoT}: \\
This starts with \textsc{\textbf{Instruction}} followed by \textsc{\textbf{Prompt with example(s)}},  \textsc{\textbf{Reasoning}}, \textsc{\textbf{Prompt+}} and finally \textsc{\textbf{System role}} and \textsc{\textbf{Response}}. \\

\hrule
\vspace{0.2cm}
\textbf{PaLM 2 setup follows:}
Using the context from "setup piece and configuration" \\
\textbf{Zero-shot:} \\
This starts with \textbf{\textsc{System role}} called \textsc{\textbf{Additional Instruction}} in PaLM 2, followed by \textsc{\textbf{Prompt}} and finally \textsc{\textbf{Prompt+}} and \textsc{\textbf{Response}}.\\
\textbf{Few-shot:} \\
This starts with \textbf{\textsc{System role}} called \textsc{\textbf{Additional Instruction}} followed by \textsc{\textbf{Instruction}}, \textsc{\textbf{Prompt with example(s)}}, and finally \textsc{\textbf{Prompt+}} and \textsc{\textbf{Response}}. \\
\textbf{Few-shot CoT}: \\
This starts with \textsc{\textbf{Instruction}} followed by \textsc{\textbf{Prompt with example(s)}},  \textsc{\textbf{Reasoning}},  \textsc{\textbf{Prompt+}} and finally \textsc{\textbf{System role}} and \textsc{\textbf{Response}}. \\
\hrule
\vspace{0.2cm}
\textbf{MPT Instruct setup follows:}
Using the context from "setup piece and configuration" \\
\textbf{Zero-shot:} \\
This starts with \textbf{\textsc{System role}} also called \textsc{\textbf{Instruction}} in MPT Instruct, followed by \textsc{\textbf{Prompt}} and finally \textsc{\textbf{Prompt+}} and \textsc{\textbf{Response}}.\\
\textbf{Few-shot:} \\
This starts with \textbf{\textsc{System role}} called \textsc{\textbf{ Instruction}}, followed by \textsc{\textbf{Prompt with example(s)}}, and finally \textsc{\textbf{Prompt+}} and \textsc{\textbf{Response}}. \\
\textbf{Few-shot CoT}: \\
This starts with \textsc{\textbf{Instruction}}, followed by \textsc{\textbf{Prompt with example(s)}}, \textsc{\textbf{Reasoning}}, \textsc{\textbf{Prompt+}} and finally \textsc{\textbf{System role}}.
 \end{framed}
 \caption{LLM Setup and Configuration. }
 \label{fig:llm_config}
\end{figure*}

\begin{table*}[htb]
\subsection{Prompt Description}
\centering
\footnotesize
\begin{tabular}{|p{1cm}|p{14cm}|}
\hline
\textbf{Title} & \textbf{Prompt style based on LLM setup} \\ \hline
Simple Prompt & In the context of this sentence: The predecessor **Mississippi Power Company** was incorporated under the laws of the State of Maine on November 24, 1924 and was admitted to do business in Mississippi on \_\_December 23, 1924\_\_ and in Alabama on December 7, 1962 . Note the location of the Mississippi Power Company and December 23, 1924 as highlighted to help determine the relation given the listed options below. Please choose the MOST appropriate relation from the following options: 1. Mississippi Power Company is/was acquired on December 23, 1924. 2. Mississippi Power Company is/was formed on December 23, 1924. 3. no/other relation between Mississippi Power Company and December 23, 1924. \\ \hline
Full Instruction Prompt & Select date of formation relationship described in one sentence. Given a single sentence: The predecessor **Mississippi Power Company** was incorporated under the laws of the State of Maine on November 24, 1924 and was admitted to do business in Mississippi on \_\_December 23, 1924\_\_ and in Alabama on December 7, 1962. With 2 highlighted phrases:Mississippi Power Company and December 23, 1924, select a multiple choice answer from options below, which best describes the relation between Mississippi Power Company and December 23, 1924. Please choose the MOST appropriate relation from the following options: 1. Mississippi Power Company is/was formed on December 23, 1924. 2. Mississippi Power Company is/was acquired on December 23, 1924. 3. no/other relation between Mississippi Power Company and December 23, 1924. \\ \hline 
1-Shot Prompt & Select the statement that best describes the relation in the example sentence below. Ignore any grammatical errors. If there are multiple options, please choose the one that is clearest and most obvious from the sentence. \textbackslash{}n\textbackslash{}nExample Sentence 1:**LecTec** was organized in 1977 as a Minnesota corporation and went public in \_\_December 1986\_\_. \textbackslash{}n Answer to Example 1: LecTec was formed/incorporated on/in December 1986. \textbackslash{}n Following the example above, read through this sentence: The predecessor **Mississippi Power Company** was incorporated under the laws of the State of Maine on November 24, 1924 and was admitted to do business in Mississippi on \_\_December 23, 1924\_\_ and in Alabama on December 7, 1962 . Given the location of the Mississippi Power Company and December 23, 1924 as highlighted, choose an answer from listed options below. \textbackslash{}n Please choose the MOST appropriate relation from the following options: \textbackslash{}n 1. Mississippi Power Company is/was acquired on December 23, 1924\textbackslash{}n 2. Mississippi Power Company is/was formed on December 23, 1924\textbackslash{}n 3. no/other relation between Mississippi Power Company and December 23, 1924.  \\ \hline 
5-Shot Prompt & Select the statement that best describes the relation in the example sentence below. Ignore any grammatical errors. If there are multiple options, please choose the one that is clearest and most obvious from the sentence. \textbackslash{}n\textbackslash{}n Example Sentence 1:**LecTec** was organized in 1977 as a Minnesota corporation and went public in \_\_December 1986\_\_. \textbackslash{}n Answer to Example 1: LecTec was formed/incorporated on/in December 1986. \textbackslash{}n Example Sentence 2: The assets of **Unified Payments , LLC** were acquired by us in \_\_April 2013\_\_.\textbackslash{}n Answer to Example 2: Unified Payments, LLC was acquired in April 2013. \textbackslash{}n Example Sentence 3: Since \_\_July 6, 2016\_\_ , Pinnacle West has issued four parental guarantees for 4CA relating to payment obligations arising from 4CA s acquisition of El Paso s 7 \% interest in **Four Corners** , and pursuant to the Four Corners participation agreement payment obligations arising from 4CA s ownership interest in Four Corners. \textbackslash{}n Answer to Example 3: No relation between Four Corners and July 6 , 2016. \textbackslash{}n Example Sentence 4: In\_\_ 2014\_\_ , \$ 148 million cash proceeds , net of cash sold , from Sempra Renewables sale of 50 - percent equity interests in **Copper Mountain Solar 3** ( \$ 66 million ) and Broken Bow 2 Wind ( \$ 58 million ) , and Sempra Mexico s sale of a 50 - percent equity interest in Energ a Sierra Ju rez ( \$ 24 million ) ; and .\textbackslash{}n Answer to Example 4: No relation between Copper Mountain Solar 3 and 2014.\textbackslash{}n Example Sentence 5: **Zendex** was incorporated in the state of Utah in \_\_March 2011\_\_ to create an online platform for the sale of art .\textbackslash{}n Answer to Example 5:Zendex was formed in March 2011. \textbackslash{}n\textbackslash{}n Following the example above, read through this sentence: The predecessor **Mississippi Power Company** was incorporated under the laws of the State of Maine on November 24, 1924 and was admitted to do business in Mississippi on \_\_December 23, 1924\_\_ and in Alabama on December 7, 1962 . Given the location of the Mississippi Power Company and December 23, 1924 as highlighted, choose an answer from listed options below. \textbackslash{}n Please choose the MOST appropriate relation from the following options: \textbackslash{}n 1. Mississippi Power Company is/was formed on December 23, 1924\textbackslash{}n 2. Mississippi Power Company is/was acquired on/in December 23, 1924\textbackslash{}n 3. no/other relation between Mississippi Power Company and December 23, 1924. \\ \hline
\end{tabular}
\end{table*}

\begin{table*}[h!]
{\footnotesize
\begin{tabular}{|p{2cm}|p{13cm}|}
\hline
\textbf{Title} & \textbf{Prompt style based on LLM setup} \\ \hline
1-Shot CoT Prompt & Select the statement that best describes the relation in the example sentence below. Ignore any grammatical errors. If there are multiple options, please choose the one that is clearest and most obvious from the sentence. \textbackslash{}n\textbackslash{}n Example Sentence 1:**LecTec** was organized in \_\_1977\_\_ as a Minnesota corporation and went public in December 1986. \textbackslash{}n Answer to Example 1: LecTec was formed/incorporated on/in 1977. \textbackslash{}n The reasoning for the above answer is that the highlighted portion of the question, LecTec, corresponds with the entity being discussed, and the year 1977 refers to when LecTec was organized or incorporated, both of which are accurately reflected in the answer.\textbackslash{}n Following the example above, read through this sentence: The predecessor **Mississippi Power Company** was incorporated under the laws of the State of Maine on November 24, 1924 and was admitted to do business in Mississippi on \_\_December 23, 1924\_\_ and in Alabama on December 7, 1962 . Given the location of the Mississippi Power Company and December 23, 1924 as highlighted, choose an answer from listed options below. \textbackslash{}n Please choose the MOST appropriate relation from the following options: \textbackslash{}n 1. Mississippi Power Company is/was acquired on December 23, 1924\textbackslash{}n 2. Mississippi Power Company is/was formed on December 23, 1924\textbackslash{}n 3. no/other relation between Mississippi Power Company and December 23, 1924.  \\ \cline{1-2} 
5-Shot CoT Prompt & Select the statement that best describes the relation in the example sentence below. Ignore any grammatical errors. If there are multiple options, please choose the one that is clearest and most obvious from the sentence. \textbackslash{}n\textbackslash{}n Example Sentence 1:**LecTec** was organized in \_\_1977\_\_ as a Minnesota corporation and went public in December 1986. \textbackslash{}n Answer to Example 1: LecTec was formed/incorporated on/in 1977. \textbackslash{}n The reasoning for the above answer is that the highlighted portion of the question, LecTec, corresponds with the entity being discussed, and the year 1977 refers to when LecTec was organized or incorporated, both of which are accurately reflected in the answer. \textbackslash{}n Example Sentence 2: The assets of **Unified Payments , LLC** were acquired by us in \_\_April 2013\_\_.\textbackslash{}n Answer to Example 2: Unified Payments, LLC was acquired in April 2013. \textbackslash{}n The reasoning for the answer above is that the highlighted portions of the question indicate the key elements of the event being asked about: Unified Payments, LLC being the entity that was acquired and April 2013 being the time when the acquisition took place, both of which are directly stated in the answer. \textbackslash{}n Example Sentence 3: Since \_\_July 6, 2016\_\_ , Pinnacle West has issued four parental guarantees for 4CA relating to payment obligations arising from 4CA s acquisition of El Paso s 7 \% interest in **Four Corners** , and pursuant to the Four Corners participation agreement payment obligations arising from 4CA s ownership interest in Four Corners. \textbackslash{}n Answer to Example 3: No relation between Four Corners and July 6, 2016. \textbackslash{}n We are only interested in identifying if the organization mentioned was formed on the specified date or acquired by another organization on the specified date.  Since Four Corners was neither formed on  July 6, 2016 nor acquired by another company on  July 6, 2016,  there is no relation between Four Corners and July 6, 2016.\textbackslash{}n Example Sentence 4: In\_\_ 2014\_\_ , \$ 148 million cash proceeds , net of cash sold , from Sempra Renewables sale of 50 - percent equity interests in **Copper Mountain Solar 3** ( \$ 66 million ) and Broken Bow 2 Wind ( \$ 58 million ) , and Sempra Mexico s sale of a 50 - percent equity interest in Energ a Sierra Ju rez ( \$ 24 million ) ; and .\textbackslash{}n Answer to Example 4: No relation between Copper Mountain Solar 3 and 2014.\textbackslash{}n We are only interested in identifying if the organization mentioned was formed on the specified date or acquired by another organization on the specified date.  Since Copper Mountain Solar 3 was neither formed in 2014 nor acquired by another company in 2014,  there is no relation between Copper Mountain Solar 3 and 2014. \textbackslash{}n Example Sentence 5: **Zendex** was incorporated in the state of Utah in \_\_March 2011\_\_ to create an online platform for the sale of art .\textbackslash{}n Answer to Example 5:Zendex was formed in March 2011. \textbackslash{}n\textbackslash{}n The incorporation of Zendex in March 2011 suggests that this is the official date when the company was legally established and recognized as a corporate entity in the state of Utah. Hence Zendex was formed on March 2011. \textbackslash{}n Following the example above, read through this sentence: The predecessor **Mississippi Power Company** was incorporated under the laws of the State of Maine on November 24, 1924 and was admitted to do business in Mississippi on \_\_December 23, 1924\_\_ and in Alabama on December 7, 1962 . Given the location of the Mississippi Power Company and December 23, 1924 as highlighted, choose an answer from listed options below. \textbackslash{}n Please choose the MOST appropriate relation from the following options: \textbackslash{}n 1. Mississippi Power Company is/was formed on December 23, 1924\textbackslash{}n 2. Mississippi Power Company is/was acquired on/in December 23, 1924\textbackslash{}n 3. no/other relation between Mississippi Power Company and December 23, 1924. \\ \hline
\end{tabular}
\caption{Prompts for Entity-Pair: ORG-DATE}
}
\end{table*}

\clearpage
\begin{table}[htb]
\subsection{Metrics for LLM Annotators}
\centering
\resizebox{\textwidth}{!}{
\begin{tabular}{cclccccccc}
\hline
\multicolumn{10}{c}{\textbf{RUN 1: Micro F1 Score / Accuracy(\%)}}                                                                                                                                                                                                                                                                                                                                                \\ \hline
\multicolumn{1}{c}{\multirow{2}{*}{\textbf{LLM}}} & \multicolumn{1}{c|}{\multirow{2}{*}{\textbf{Annotator}}} & \multicolumn{1}{c|}{\multirow{2}{*}{\textbf{\begin{tabular}[c]{@{}c@{}}Annotator \\ Description\end{tabular}}}} & \multicolumn{6}{c|}{\textbf{Entity-Pair}}                                                                                                      & \multirow{2}{*}{\textbf{Total}} \\ \cline{4-9}
\multicolumn{1}{c}{}                              & \multicolumn{1}{c|}{}                                    & \multicolumn{1}{c|}{}                                                                                           & \textbf{ORG-GPE}   & \textbf{ORG-ORG} & \textbf{ORG-DATE}  & \textbf{ORG-MONEY} & \textbf{PER-ORG}   & \multicolumn{1}{c|}{\textbf{PER-TITLE}} &                                 \\ \hline
\multicolumn{1}{c}{\multirow{12}{*}{GPT-4}}       & annotator1                                               & \multicolumn{1}{l|}{simple prompt, temp=0.2}                                                                    & 80.1/74.8          & 15.4/38.4        & 48.9/67.0          & 48.4/43.4          & 70.8/67.8          & \multicolumn{1}{c|}{92.7/86.9}          & 67.2/63.2                       \\
\multicolumn{1}{c}{}                              & annotator2                                               & \multicolumn{1}{l|}{simple prompt, temp=0.7}                                                                    & 80.3/74.6          & 15.1/37.1        & 51.6/70.0          & \textbf{48.9/44.5} & 72.1/69.3          & \multicolumn{1}{c|}{93.2/87.6}          & 67.8/63.7                       \\
\multicolumn{1}{c}{}                              & annotator3                                               & \multicolumn{1}{l|}{full instruction, temp=0.2}                                                                 & 80.9/75.8          & 15.7/36.7        & 56.3/74.5          & 47.8/43.1          & \textbf{72.8/69.9} & \multicolumn{1}{c|}{93.7/88.7}          & \textbf{68.6/64.7}              \\
\multicolumn{1}{c}{}                              & annotator4                                               & \multicolumn{1}{l|}{full instruction, temp=0.7}                                                                 & \textbf{80.9/75.8} & 15.5/36.9        & 55.4/73.8          & 47.6/42.7          & 71.8/69.3          & \multicolumn{1}{c|}{93.5/88.2}          & 68.2/64.4                       \\
\multicolumn{1}{c}{}                              & annotator5                                               & \multicolumn{1}{l|}{1-shot, temp=0.2}                                                                           & 79.6/73.9          & 15.4/30.1        & 48.0/65.7          & 47.4/42.0          & 62.5/60.0          & \multicolumn{1}{c|}{94.4/89.9}          & 65.0/60.1                       \\
\multicolumn{1}{c}{}                              & annotator6                                               & \multicolumn{1}{l|}{1-shot, temp=0.7}                                                                           & 79.8/73.9          & 14.5/28.9        & 50.3/68.1          & 48.0/42.3          & 62.5/60.0          & \multicolumn{1}{c|}{94.3/89.8}          & 65.1/60.1                       \\
\multicolumn{1}{c}{}                              & annotator7                                               & \multicolumn{1}{l|}{5-shot, temp=0.2}                                                                           & 79.3/73.9          & 15.7/35.5        & 59.3/78.0          & 48.0/41.3          & 71.1/67.8          & \multicolumn{1}{c|}{93.3/87.9}          & 67.8/64.0                       \\
\multicolumn{1}{c}{}                              & annotator8                                               & \multicolumn{1}{l|}{5-shot, temp=0.7}                                                                           & 78.9/73.5          & 15.0/35.3        & 58.9/77.6          & 47.1/40.2          & 72.0/69.1          & \multicolumn{1}{c|}{93.2/87.8}          & 67.6/63.8                       \\
\multicolumn{1}{c}{}                              & annotator9                                               & \multicolumn{1}{l|}{COT 1-shot, temp=0.2}                                                                       & 79.6/74.1          & 16.1/32.5        & 36.8/47.3          & 48.9/43.4          & 63.7/61.0          & \multicolumn{1}{c|}{94.4/89.8}          & 64.3/58.1                       \\
\multicolumn{1}{c}{}                              & annotator10                                              & \multicolumn{1}{l|}{COT 1-shot, temp=0.7}                                                                       & 80.2/74.6          & 16.2/32.7        & 37.8/48.9          & 47.5/41.3          & 63.7/60.8          & \multicolumn{1}{c|}{\textbf{94.7/90.2}} & 64.6/58.4                       \\
\multicolumn{1}{c}{}                              & annotator11                                              & \multicolumn{1}{l|}{COT 5-shot, temp=0.2}                                                                       & 79.4/73.7          & 16.2/37.8        & \textbf{65.4/83.2} & 46.3/42.7          & 70.6/67.6          & \multicolumn{1}{c|}{92.8/87.0}          & 68.2/65.2                       \\
\multicolumn{1}{c}{}                              & annotator12                                              & \multicolumn{1}{l|}{COT 5-shot, temp=0.7}                                                                       & 79.6/73.8          & \textbf{17.0/38.4 }       & 65.2/83.0          & 46.0/43.1          & 70.9/67.8          & \multicolumn{1}{c|}{92.9/87.0}          & 68.4/65.5                       \\ \hline
\multirow{12}{*}{PaLM 2}                          & annotator1                                               & \multicolumn{1}{l|}{simple prompt, temp=0.2}                                                                    & 81.0/76.9          & \textbf{13.5}/14.7        & 50.1/67.0          & 43.5/29.2          & 68.3/62.9          & \multicolumn{1}{c|}{87.2/78.5}          & 62.6/54.3                       \\
                                                  & annotator2                                               & \multicolumn{1}{l|}{simple prompt, temp=0.7}                                                                    & 80.0/76.1          & 13.5/13.3        & 49.3/65.9          & 43.7/29.9          & 69.0/63.5          & \multicolumn{1}{c|}{93.8/90.4}          & 64.4/55.9                       \\
                                                  & annotator3                                               & \multicolumn{1}{l|}{full instruction, temp=0.2}                                                                 & 79.3/75.5          & 13.2/14.0        & 49.3/66.2          & 44.2/31.3          & 67.7/62.1          & \multicolumn{1}{c|}{87.0/77.9}          & 61.9/53.6                       \\
                                                  & annotator4                                               & \multicolumn{1}{l|}{full instruction, temp=0.7}                                                                 & 80.5/76.5          & 13.2/12.5        & 49.9/66.6          & 43.7/29.9          & 68.0/62.7          & \multicolumn{1}{c|}{94.1/90.7}          & 64.3/55.8                       \\
                                                  & annotator5                                               & \multicolumn{1}{l|}{1-shot, temp=0.2}                                                                           & 86.4/81.4          & 13.0/\textbf{33.0}        & 48.7/64.6          & 42.7/29.2          & 67.5/63.7          & \multicolumn{1}{c|}{90.9/84.0}          & 66.7/60.5                       \\
                                                  & annotator6                                               & \multicolumn{1}{l|}{1-shot, temp=0.7}                                                                           & \textbf{87.1/82.1} & 13.0/22.1        & 57.8/77.6          & 42.2/31.0          & 64.2/60.4          & \multicolumn{1}{c|}{\textbf{95.9/93.0}} & 67.5/61.3                       \\
                                                  & annotator7                                               & \multicolumn{1}{l|}{5-shot, temp=0.2}                                                                           & 81.3/74.9          & 12.4/20.2        & 55.4/73.3          & 41.4/31.3          & 70.6/67.2          & \multicolumn{1}{c|}{95.2/91.9}          & 66.1/59.4                       \\
                                                  & annotator8                                               & \multicolumn{1}{l|}{5-shot, temp=0.7}                                                                           & 86.5/82.3          & 12.6/27.2        & \textbf{63.8/81.8}          & \textbf{44.6/35.9} & 68.4/64.9          & \multicolumn{1}{c|}{95.0/91.5}          & \textbf{69.0/63.9}              \\
                                                  & annotator9                                               & \multicolumn{1}{l|}{COT 1-shot, temp=0.2}                                                                       & 84.7/81.0          & 12.4/12.7        & 46.7/63.7          & 43.2/28.5          & 67.2/63.3          & \multicolumn{1}{c|}{93.2/87.6}          & 64.5/55.7                       \\
                                                  & annotator10                                              & \multicolumn{1}{l|}{COT 1-shot, temp=0.7}                                                                       & 84.4/80.0          & 12.2/16.9        & 49.0/68.6          & 40.8/29.5          & 65.3/61.0          & \multicolumn{1}{c|}{93.5/88.9}          & 64.9/57.3                       \\
                                                  & annotator11                                              & \multicolumn{1}{l|}{COT 5-shot, temp=0.2}                                                                       & 81.6/77.7          & 13.4/11.4        & 50.4/66.6          & 40.7/32.4          & \textbf{71.4/67.6} & \multicolumn{1}{c|}{95.5/92.4}          & 65.4/56.9                       \\
                                                  & annotator12                                              & \multicolumn{1}{l|}{COT 5-shot, temp=0.7}                                                                       & 83.5/79.3          & 12.6/16.0        & 54.2/73.5          & 41.6/34.2          & 70.0/66.6          & \multicolumn{1}{c|}{93.2/89.6}          & 65.8/59.0                       \\ \hline
\multirow{12}{*}{MPT Instruct}                    & annotator1                                               & \multicolumn{1}{l|}{simple prompt, temp=0.2}                                                                    & 16.7/16.2          & 6.2/14.9         & 18.4/31.0          & 25.0/27.4          & 40.2/37.5          & \multicolumn{1}{c|}{17.1/15.4}          & 19.9/21.8                       \\
                                                  & annotator2                                               & \multicolumn{1}{l|}{simple prompt, temp=0.7}                                                                    & 25.3/23.1          & 5.8/20.8         & 16.5/35.7          & 13.1/29.2          & 31.3/28.5          & \multicolumn{1}{c|}{23.5/20.3}          & 20.9/25.2                       \\
                                                  & annotator3                                               & \multicolumn{1}{l|}{full instruction, temp=0.2}                                                                 & 39.2/34.8          & 5.4/15.2         & \textbf{24.2}/24.7          & \textbf{34.0}/30.2          & 31.8/30.1          & \multicolumn{1}{c|}{41.8/35.1}          & 30.3/27.3                       \\
                                                  & annotator4                                               & \multicolumn{1}{l|}{full instruction, temp=0.7}                                                                 & 31.9/28.6          & 5.3/\textbf{24.9}         & 19.6/29.2          & 22.8/\textbf{31.0}          & 31.0/29.1          & \multicolumn{1}{c|}{32.4/27.6}          & 25.4/27.8                       \\
                                                  & annotator5                                               & \multicolumn{1}{l|}{1-shot, temp=0.2}                                                                           & 25.7/24.1          & 5.9/7.2          & 21.4/29.8          & 29.4/22.8          & 16.2/15.3          & \multicolumn{1}{c|}{17.5/16.5}          & 18.3/18.0                       \\
                                                  & annotator6                                               & \multicolumn{1}{l|}{1-shot, temp=0.7}                                                                           & 27.6/25.2          & 5.5/17.0         & 13.0/28.0          & 23.5/23.5          & 22.1/21.0          & \multicolumn{1}{c|}{36.1/31.5}          & 22.9/24.0                       \\
                                                  & annotator7                                               & \multicolumn{1}{l|}{5-shot, temp=0.2}                                                                           & 49.5/45.9          & 4.4/9.4          & 23.1/\textbf{43.5}          & 20.5/15.7          & 54.3/50.1          & \multicolumn{1}{c|}{\textbf{69.5}/56.9}          & 41.6/\textbf{36.5}                       \\
                                                  & annotator8                                               & \multicolumn{1}{l|}{5-shot, temp=0.7}                                                                           & 37.2/33.7          & 5.0/24.4         & 13.2/35.9          & 16.5/18.5          & 37.6/33.8          & \multicolumn{1}{c|}{46.1/36.0}          & 29.8/30.9                       \\
                                                  & annotator9                                               & \multicolumn{1}{l|}{COT 1-shot, temp=0.2}                                                                       & 22.7/21.4          & 6.5/6.2          & 17.5/23.5          & 29.8/21.7          & 16.1/15.1          & \multicolumn{1}{c|}{30.1/27.6}          & 20.0/18.2                       \\
                                                  & annotator10                                              & \multicolumn{1}{l|}{COT 1-shot, temp=0.7}                                                                       & 25.5/23.0          & \textbf{8.5}/18.8         & 14.0/28.0          & 19.3/19.9          & 18.3/16.9          & \multicolumn{1}{c|}{37.7/33.0}          & 22.3/23.5                       \\
                                                  & annotator11                                              & \multicolumn{1}{l|}{COT 5-shot, temp=0.2}                                                                       & \textbf{58.7/55.1} & 6.5/7.4          & 23.6/22.4          & 21.6/14.6          & \textbf{64.1/59.0}          & \multicolumn{1}{c|}{67.8/\textbf{58.8}}          & \textbf{45.2}/36.0                       \\
                                                  & annotator12                                              & \multicolumn{1}{l|}{COT 5-shot, temp=0.7}                                                                       & 41.9/37.7          & 5.6/18.8         & 22.3/27.3          & 11.8/13.9          & 49.1/44.1          & \multicolumn{1}{c|}{47.7/39.7}          & 33.7/30.7                       \\ \hline
\end{tabular}}
\vspace{0.2cm}
\parbox{0.95\textwidth}{
  \caption{First Run LLM Annotators: Micro-Averaged F1 Score/Accuracy}
  \label{llm_metric}
}
\end{table}

\begin{table*}[hbt!]
\centering
\resizebox{\textwidth}{!}{
\begin{tabular}{cclcccccccccc}
\hline
\multicolumn{10}{c}{\textbf{RUN 2: Micro F1 Score/ Accuracy(\%)}}                                                                                                                                                                                                                                                                                                                                                      \\ \hline
\multicolumn{1}{c}{\multirow{2}{*}{\textbf{LLM}}} & \multicolumn{1}{c|}{\multirow{2}{*}{\textbf{Annotator}}} & \multicolumn{1}{c|}{\multirow{2}{*}{\textbf{\begin{tabular}[c]{@{}c@{}}Annotator \\ Description\end{tabular}}}} & \multicolumn{6}{c|}{\textbf{Entity-Pair}}                                                                                                      & \multirow{2}{*}{\textbf{Total}} \\ \cline{4-9}
\multicolumn{1}{c}{}                              & \multicolumn{1}{c|}{}                                    & \multicolumn{1}{c|}{}                                                                                           & \textbf{ORG-GPE}   & \textbf{ORG-ORG} & \textbf{ORG-DATE}  & \textbf{ORG-MONEY} & \textbf{PER-ORG}   & \multicolumn{1}{c|}{\textbf{PER-TITLE}} &                                 \\ \hline
\multicolumn{1}{c}{\multirow{12}{*}{GPT-4}}       & annotator1                                               & \multicolumn{1}{l|}{simple prompt, temp=0.2}                                                                    & 80.5/75.2          & 15.3/38.0        & 50.6/68.4          & 48.6/\textbf{44.8}          & 71.1/68.0          & \multicolumn{1}{c|}{92.9/87.2}          & 67.6/63.6                       \\
\multicolumn{1}{c}{}                              & annotator2                                               & \multicolumn{1}{l|}{simple prompt, temp=0.7}                                                                    & 80.3/74.9          & 15.0/38.3        & 49.9/67.9          & \textbf{48.7}/44.1          & 71.0/68.2          & \multicolumn{1}{c|}{92.7/86.9}          & 67.4/63.4                       \\
\multicolumn{1}{c}{}                              & annotator3                                               & \multicolumn{1}{l|}{full instruction, temp=0.2}                                                                 & \textbf{81.3/75.9} & 15.4/36.6        & 55.9/74.4          & 47.4/42.0          & 72.3/69.5          & \multicolumn{1}{c|}{93.5/88.4}          & 68.4/64.5                       \\
\multicolumn{1}{c}{}                              & annotator4                                               & \multicolumn{1}{l|}{full instruction, temp=0.7}                                                                 & 80.9/75.8          & 15.4/37.5        & 57.7/75.6          & 47.6/42.7          & \textbf{72.4/69.7} & \multicolumn{1}{c|}{93.1/87.8}          & 68.5/64.8                       \\
\multicolumn{1}{c}{}                              & annotator5                                               & \multicolumn{1}{l|}{1-shot, temp=0.2}                                                                           & 79.9/74.5          & 14.3/29.4        & 48.8/66.4          & 47.4/42.0          & 62.8/60.2          & \multicolumn{1}{c|}{94.2/89.6}          & 65.0/60.1                       \\
\multicolumn{1}{c}{}                              & annotator6                                               & \multicolumn{1}{l|}{1-shot, temp=0.7}                                                                           & 79.1/73.4          & 14.3/29.1        & 48.2/65.7          & 48.0/42.3          & 62.5/60.0          & \multicolumn{1}{c|}{\textbf{94.7/90.4}} & 64.8/59.8                       \\
\multicolumn{1}{c}{}                              & annotator7                                               & \multicolumn{1}{l|}{5-shot, temp=0.2}                                                                           & 78.5/73.5          & 15.9/34.9        & 58.2/76.9          & 47.6/40.2          & 71.1/67.8          & \multicolumn{1}{c|}{93.4/88.1}          & 67.4/63.5                       \\
\multicolumn{1}{c}{}                              & annotator8                                               & \multicolumn{1}{l|}{5-shot, temp=0.7}                                                                           & 79.7/74.4          & 15.0/35.2        & 58.9/77.6          & 48.0/41.3          & 71.7/68.2          & \multicolumn{1}{c|}{93.2/87.8}          & 67.8/64.0                       \\
\multicolumn{1}{c}{}                              & annotator9                                               & \multicolumn{1}{l|}{COT 1-shot, temp=0.2}                                                                       & 80.4/74.8          & 16.4/32.9        & 37.3/48.0          & 48.3/43.1          & 64.7/61.9          & \multicolumn{1}{c|}{94.5/89.9}          & 64.7/58.6                       \\
\multicolumn{1}{c}{}                              & annotator10                                              & \multicolumn{1}{l|}{COT 1-shot, temp=0.7}                                                                       & 80.2/74.8          & 16.3/32.2        & 36.4/46.9          & 47.2/41.6          & 65.5/62.9          & \multicolumn{1}{c|}{94.5/89.9}          & 64.6/58.3                       \\
\multicolumn{1}{c}{}                              & annotator11                                              & \multicolumn{1}{l|}{COT 5-shot, temp=0.2}                                                                       & 79.9/74.5          & \textbf{16.6}/37.7        & \textbf{65.2/83.0} & 47.0/43.1          & 71.3/68.2          & \multicolumn{1}{c|}{92.9/87.0}          & \textbf{68.5/65.5}              \\
\multicolumn{1}{c}{}                              & annotator12                                              & \multicolumn{1}{l|}{COT 5-shot, temp=0.7}                                                                       & 80.2/74.5          & 16.5/\textbf{37.9}        & 64.9/82.9          & 46.8/42.7          & 70.7/67.4          & \multicolumn{1}{c|}{92.9/87.0}          & 68.4/65.3                       \\ \hline
\multirow{12}{*}{PaLM 2}                          & annotator1                                               & \multicolumn{1}{l|}{simple prompt, temp=0.2}                                                                    & 80.1/75.9          & 14.0/14.1        & 49.7/67.0          & 43.5/29.2          & 66.9/60.8          & \multicolumn{1}{c|}{87.0/77.7}          & 62.0/53.5                       \\
                                                  & annotator2                                               & \multicolumn{1}{l|}{simple prompt, temp=0.7}                                                                    & 80.6/76.6          & 13.8/13.8        & 48.4/65.3          & \textbf{43.9}/30.6          & 67.8/61.9          & \multicolumn{1}{c|}{94.1/91.1}          & 64.5/56.0                       \\
                                                  & annotator3                                               & \multicolumn{1}{l|}{full instruction, temp=0.2}                                                                 & 80.2/76.3          & 13.6/13.8        & 49.6/66.2          & \textbf{43.9}/30.6          & 69.1/63.7          & \multicolumn{1}{c|}{87.1/78.0}          & 62.4/53.9                       \\
                                                  & annotator4                                               & \multicolumn{1}{l|}{full instruction, temp=0.7}                                                                 & 79.9/75.8          & \textbf{14.1}/14.7        & 49.5/66.1          & 43.6/29.5          & 68.5/63.1          & \multicolumn{1}{c|}{94.1/91.0}          & 64.5/56.2                       \\
                                                  & annotator5                                               & \multicolumn{1}{l|}{1-shot, temp=0.2}                                                                           & 86.1/81.0          & 13.1/31.7        & 47.3/63.0          & 42.5/28.5          & 67.2/63.3          & \multicolumn{1}{c|}{90.3/83.1}          & 66.1/59.6                       \\
                                                  & annotator6                                               & \multicolumn{1}{l|}{1-shot, temp=0.7}                                                                           & \textbf{87.4/82.0} & 13.1/21.8        & 55.3/74.7          & 42.2/29.9          & 63.5/60.0          & \multicolumn{1}{c|}{\textbf{95.7/92.2}} & 67.1/60.4                       \\
                                                  & annotator7                                               & \multicolumn{1}{l|}{5-shot, temp=0.2}                                                                           & 82.1/75.8          & 11.8/18.3        & 56.1/74.0          & 41.6/32.7          & 69.6/66.4          & \multicolumn{1}{c|}{94.4/90.5}          & 65.8/59.0                       \\
                                                  & annotator8                                               & \multicolumn{1}{l|}{5-shot, temp=0.7}                                                                           & 86.0/81.8          & 13.3/\textbf{27.3}        & \textbf{64.3/82.1} & 43.2/\textbf{35.9}          & 68.5/65.4          & \multicolumn{1}{c|}{93.6/89.3}          & \textbf{68.4/63.6}              \\
                                                  & annotator9                                               & \multicolumn{1}{l|}{COT 1-shot, temp=0.2}                                                                       & 84.7/81.0          & 12.5/12.8        & 47.4/65.2          & 43.3/28.8          & 68.7/64.7          & \multicolumn{1}{c|}{92.7/87.0}          & 64.8/56.1                       \\
                                                  & annotator10                                              & \multicolumn{1}{l|}{COT 1-shot, temp=0.7}                                                                       & 84.6/80.1          & 11.7/16.3        & 49.1/68.6          & 41.1/29.2          & 65.5/61.9          & \multicolumn{1}{c|}{94.0/89.8}          & 64.9/57.5                       \\
                                                  & annotator11                                              & \multicolumn{1}{l|}{COT 5-shot, temp=0.2}                                                                       & 82.4/78.6          & 13.3/11.8        & 50.7/67.0          & 43.2/35.2          & \textbf{71.5/67.8} & \multicolumn{1}{c|}{95.1/92.1}          & 65.8/57.5                       \\
                                                  & annotator12                                              & \multicolumn{1}{l|}{COT 5-shot, temp=0.7}                                                                       & 82.4/78.2          & \textbf{14.1}/17.6        & 54.5/74.2          & 41.5/32.4          & 69.0/65.6          & \multicolumn{1}{c|}{94.6/91.6}          & 66.1/59.4                       \\ \hline
\multirow{12}{*}{MPT Instruct}                    & annotator1                                               & \multicolumn{1}{l|}{simple prompt, temp=0.2}                                                                    & 17.7/17.0          & 6.5/15.8         & 17.7/31.2          & 25.2/26.0          & 40.4/37.5          & \multicolumn{1}{c|}{16.7/14.7}          & 20.1/21.9                       \\
                                                  & annotator2                                               & \multicolumn{1}{l|}{simple prompt, temp=0.7}                                                                    & 21.0/20.3          & 6.4/20.0         & 15.1/34.7          & 15.1/22.4          & 33.2/30.5          & \multicolumn{1}{c|}{25.5/21.7}          & 20.6/24.2                       \\
                                                  & annotator3                                               & \multicolumn{1}{l|}{full instruction, temp=0.2}                                                                 & 42.2/38.0          & 7.2/13.9         & 24.0/24.2          & 31.2/\textbf{28.5}          & 32.9/30.7          & \multicolumn{1}{c|}{44.7/37.4}          & 31.9/27.9                       \\
                                                  & annotator4                                               & \multicolumn{1}{l|}{full instruction, temp=0.7}                                                                 & 31.4/28.3          & 5.2/\textbf{23.3}         & 19.0/31.8          & 18.2/27.0          & 26.9/24.9          & \multicolumn{1}{c|}{33.0/26.9}          & 24.2/26.8                       \\
                                                  & annotator5                                               & \multicolumn{1}{l|}{1-shot, temp=0.2}                                                                           & 28.2/25.9          & 4.7/5.7          & 21.5/29.6          & \textbf{33.4}/23.1          & 16.3/15.3          & \multicolumn{1}{c|}{17.3/16.5}          & 18.9/18.0                       \\
                                                  & annotator6                                               & \multicolumn{1}{l|}{1-shot, temp=0.7}                                                                           & 28.7/26.1          & 5.3/17.5         & 16.3/29.6          & 25.6/27.8          & 19.1/19.4          & \multicolumn{1}{c|}{33.2/29.3}          & 22.4/24.3                       \\
                                                  & annotator7                                               & \multicolumn{1}{l|}{5-shot, temp=0.2}                                                                           & 54.9/51.1          & 4.5/7.3          & 22.6/38.3          & 18.8/14.6          & 55.4/52.2          & \multicolumn{1}{c|}{\textbf{71.1/59.8}} & 43.3/36.9                       \\
                                                  & annotator8                                               & \multicolumn{1}{l|}{5-shot, temp=0.7}                                                                           & 41.1/36.3          & 4.4/21.7         & 13.0/\textbf{35.9}          & 16.2/17.1          & 34.3/31.1          & \multicolumn{1}{c|}{50.2/41.7}          & 31.1/31.3                       \\
                                                  & annotator9                                               & \multicolumn{1}{l|}{COT 1-shot, temp=0.2}                                                                       & 24.1/22.4          & \textbf{7.8}/8.5          & 17.1/24.0          & 31.2/22.1          & 15.0/14.2          & \multicolumn{1}{c|}{28.4/26.3}          & 20.1/18.7                       \\
                                                  & annotator10                                              & \multicolumn{1}{l|}{COT 1-shot, temp=0.7}                                                                       & 28.3/26.6          & 7.0/16.1         & 11.5/25.3          & 19.8/21.0          & 17.0/16.9          & \multicolumn{1}{c|}{37.1/31.8}          & 22.1/22.9                       \\
                                                  & annotator11                                              & \multicolumn{1}{l|}{COT 5-shot, temp=0.2}                                                                       & \textbf{58.8/55.5} & 6.0/8.0          & \textbf{25.4}/23.5          & 20.1/13.9          & \textbf{61.2/56.3} & \multicolumn{1}{c|}{68.6/59.7}          & \textbf{45.2/36.1}              \\
                                                  & annotator12                                              & \multicolumn{1}{l|}{COT 5-shot, temp=0.7}                                                                       & 45.5/40.8          & 4.3/15.3         & 15.9/26.9          & 21.0/22.1          & 46.2/42.1          & \multicolumn{1}{c|}{48.2/40.6}          & 34.0/30.9                       \\ \hline
\end{tabular}}
\vspace{0.2cm}
\parbox{0.95\textwidth}{
  \caption{LLM Annotators: Micro-average F1-Score / Accuracy for second run}
  \label{llm_metric2}
}
\end{table*}

\begin{table}[htb]
\subsection{Statistical Tests}
\centering
\resizebox{\textwidth}{!}{
\begin{tabular}{llcc}
\hline
\multicolumn{4}{c}{\textbf{Are Difference Statistically significant at alpha = 0.05?}} \\ \hline
\multicolumn{1}{c|}{\textbf{Null Hypothesis}} & \multicolumn{1}{c|}{\textbf{LLM}} & \multicolumn{1}{c|}{\textbf{\begin{tabular}[c]{@{}c@{}}Micro-Averaged F1 Scores\\ P-values\end{tabular}}} & \textbf{\begin{tabular}[c]{@{}c@{}}Accuracy\\ P-values\end{tabular}} \\ \hline
\multicolumn{1}{c|}{\multirow{3}{*}{\begin{tabular}[c]{@{}c@{}}Ho:There is no significant difference in metric when we change temperature \\ setting\\ i.e. Ho: metrics at temp0.2 =metrics at temp0.7\end{tabular}}} & GPT-4 & 0.950 & 0.975 \\
\multicolumn{1}{c|}{} & PaLM 2 & 0.053 & 0.062 \\
\multicolumn{1}{c|}{} & MPT Instruct & 0.481 & 0.734 \\ \hline
\multicolumn{1}{c|}{\multirow{3}{*}{\begin{tabular}[c]{@{}c@{}}Ho:At temperature setting = 0.2, there is no significant difference in metric \\ when we compare run1 and run2\\ i.e. Ho: metrics at temp0.2\_first\_run =metrics at temp0.2\_second\_run\end{tabular}}} & GPT-4 & 0.935 & 0.959 \\
\multicolumn{1}{c|}{} & PaLM 2 & 0.964 & 0.932 \\
\multicolumn{1}{c|}{} & MPT Instruct & 0.921 & 0.954 \\ \hline
\multicolumn{1}{c|}{\multirow{3}{*}{\begin{tabular}[c]{@{}c@{}}Ho:At temperature setting = 0.7, there is no significant difference in metric\\  when we compare run1 and run2\\ i.e. Ho: metrics at temp0.7\_first\_run =metrics at temp0.7\_second\_run\end{tabular}}} & GPT-4 & 0.973 & 0.976 \\
\multicolumn{1}{c|}{} & PaLM 2 & 0.948 & 0.992 \\
\multicolumn{1}{c|}{} & MPT Instruct & 0.974 & 0.889 \\ \hline
\multicolumn{1}{c|}{\multirow{3}{*}{\begin{tabular}[c]{@{}c@{}}Ho:There is no significant difference when we compare average metrics across \\ first run and second for the different temperature setting\\ i.e Ho: avg\_metric at temp0.2 =avg\_metric at temp0.7\end{tabular}}} & GPT-4 & 0.967 & 0.983 \\
\multicolumn{1}{c|}{} & PaLM 2 & 0.195 & 0.213 \\
\multicolumn{1}{c|}{} & MPT Instruct & 0.493 & 0.909 \\ \hline
\multicolumn{1}{c|}{\multirow{3}{*}{\begin{tabular}[c]{@{}c@{}}Ho:There is no significant difference between LLM\\ metrics when we compare one to the other.\\ i.e Ho: LLM1 avg\_metric at temp 0.2 = LLM2 avg\_metric at temp 0.2\end{tabular}}} & GPT-4-vs-PaLM 2 & 0.055 & \textbf{0.004*} \\
\multicolumn{1}{c|}{} & GPT-4-vs-MPT Instruct & \textbf{0.000*} & \textbf{0.000*} \\
\multicolumn{1}{c|}{} & PaLM 2-vs-MPT Instruct & \textbf{0.000*} & \textbf{0.000*} \\ \hline
\end{tabular}
}

\vspace{0.2cm}
\parbox{0.95\textwidth}{
  \caption{Statistical Significance of Metric Difference: At alpha = 0.05, we test if the difference captured are statistical significant. For these hypotheses, we either reject or fail to reject the null hypothesis.}
  \label{tab:statistical_test}
}
\end{table}

\clearpage
\begin{table}[htb]
\subsection{Inter Annotator Agreement}
 \centering
 \resizebox{\textwidth}{!}{
 \begin{tabular}{l|l|ccc}
 \hline
 \multicolumn{1}{c|}{} & \multicolumn{1}{c|}{} & \multicolumn{3}{l}{\textbf{Inter annotator agreement (IAA)}} \\ \cline{3-5} 
 \multicolumn{1}{c|}{\multirow{-2}{*}{\textbf{What we measure}}} & \multicolumn{1}{c|}{\multirow{-2}{*}{\textbf{Prompt Type}}} & \multicolumn{1}{c|}{GPT-4} & \multicolumn{1}{c|}{PaLM 2} & MPT Instruct \\ \hline
 \multicolumn{1}{c|}{} & simple\_prompt, temp = 0.2 & \cellcolor[HTML]{FFFFFF}0.96 & \cellcolor[HTML]{FFFFFF}0.88 & \cellcolor[HTML]{FFFFFF}0.62 \\
 \multicolumn{1}{c|}{} & simple\_prompt, temp = 0.7 & 0.94 & 0.89 & 0.19 \\
 \multicolumn{1}{c|}{} & full\_instructn\_prompt, temp = 0.2 & 0.97 & 0.87 & 0.67 \\
 \multicolumn{1}{c|}{} & full\_instructn\_prompt, temp = 0.7 & 0.94 & 0.89 & 0.23 \\
 \multicolumn{1}{c|}{} & 1shot\_prompt, temp = 0.2 & 0.96 & 0.91 & 0.55 \\
 \multicolumn{1}{c|}{} & 1shot\_prompt, temp = 0.7 & 0.95 & 0.86 & 0.2 \\
 \multicolumn{1}{c|}{} & 5shot\_prompt, temp = 0.2 & 0.96 & 0.92 & 0.51 \\
 \multicolumn{1}{c|}{} & 5shot\_prompt, temp = 0.7 & 0.95 & 0.86 & 0.19 \\
 \multicolumn{1}{c|}{} & cot 1shot\_prompt, temp = 0.2 & 0.95 & 0.94 & 0.53 \\
 \multicolumn{1}{c|}{} & cot 1shot\_prompt, temp = 0.7 & \cellcolor[HTML]{FFFFFF}0.93 & 0.84 & 0.22 \\
 \multicolumn{1}{c|}{} & cot 5shot\_prompt, temp = 0.2 & \cellcolor[HTML]{FFFFFF}0.97 & 0.93 & 0.6 \\
 \multicolumn{1}{c|}{\multirow{-12}{*}{\begin{tabular}[l]{@{}l@{}}Same Prompt at same temperature setting\\(run twice).Using Cohen Kappa\end{tabular}}} & cot 5shot\_prompt, temp = 0.7 & \cellcolor[HTML]{FFFFFF}0.96 & 0.82 & 0.23 \\ \hline
  & simple\_prompt & \cellcolor[HTML]{FFFFFF}0.94 & 0.87 & 0.3 \\
  & full\_instructn\_prompt & \cellcolor[HTML]{FFFFFF}0.95 & \cellcolor[HTML]{FFFFFF}0.86 & \cellcolor[HTML]{FFFFFF}0.34 \\
  & 1shot\_prompt & \cellcolor[HTML]{FFFFFF}0.95 & \cellcolor[HTML]{FFFFFF}0.82 & \cellcolor[HTML]{FFFFFF}0.27 \\
  & 5shot\_prompt & \cellcolor[HTML]{FFFFFF}0.96 & \cellcolor[HTML]{FFFFFF}0.84 & \cellcolor[HTML]{FFFFFF}0.26 \\
  & cot1shot\_prompt & 0.95 & 0.85 & 0.28 \\
 \multirow{-6}{*}{\begin{tabular}[c]{@{}l@{}}Same Prompt at different temperature setting\\ (0.2 and 0.7) for only run 1. Using Cohen Kappa\end{tabular}} & cot5shot\_prompt & 0.96 & 0.83 & 0.34 \\ \hline
  & simple\_prompt & 0.95 & 0.87 & 0.34 \\
  & full\_instructn\_prompt & 0.95 & 0.87 & 0.37 \\
  & 1shot\_prompt & 0.95 & 0.84 & 0.31 \\
  & 5shot\_prompt & 0.96 & 0.86 & 0.3 \\
  & cot1shot\_prompt & 0.94 & 0.86 & 0.31 \\
 \multirow{-6}{*}{\begin{tabular}[c]{@{}l@{}}Same Prompt at different temperature \\ setting (run twice). Using Fleiss Kappa\end{tabular}} & cot5shot\_prompt & 0.96 & 0.85 & 0.36 \\ \hline
  & zero shot: simple\_vs\_full\_instruction & \cellcolor[HTML]{FFFFFF}0.87 & \cellcolor[HTML]{FFFFFF}0.88 & \cellcolor[HTML]{FFFFFF}0.39 \\
  & few shot: 1\_shot\_vs\_5\_shot & \cellcolor[HTML]{FFFFFF}0.84 & \cellcolor[HTML]{FFFFFF}0.79 & \cellcolor[HTML]{FFFFFF}0.28 \\
   \multirow{-3}{*}{\begin{tabular}[c]{@{}l@{}}Compare prompts within prompt types.\\ Using Cohen Kappa\end{tabular}} & cot few shot: cot\_1\_shot\_vs\_cot\_5\_shot & \cellcolor[HTML]{FFFFFF}0.8 & \cellcolor[HTML]{FFFFFF}0.82 & \cellcolor[HTML]{FFFFFF}0.28 \\ \hline
 \multicolumn{1}{c|}{Compare among prompt types. Using Fleiss Kappa} & \begin{tabular}[c]{@{}l@{}}simple\_vs\_full\_instruction\_1\_shot\_\\ vs\_5shot\_vs\_cot\_1\_shot\_vs\_cot\_5\_shot\end{tabular} & \cellcolor[HTML]{FFFFFF}0.83 & \cellcolor[HTML]{FFFFFF}0.79 & \cellcolor[HTML]{FFFFFF}0.31 \\ \hline
 \end{tabular}}
 
\vspace{0.2cm}
\parbox{0.95\textwidth}{
  \caption{LLM Inter Annotator Agreement: This table shows how consistent outputs from each LLMs are within and accross prompt types and within and accross different temperature settings.}
  \label{tab:LLM_IAA}
}
 \end{table}

\begin{figure*}[b!]
\centering
\includegraphics[width=0.9\textwidth]{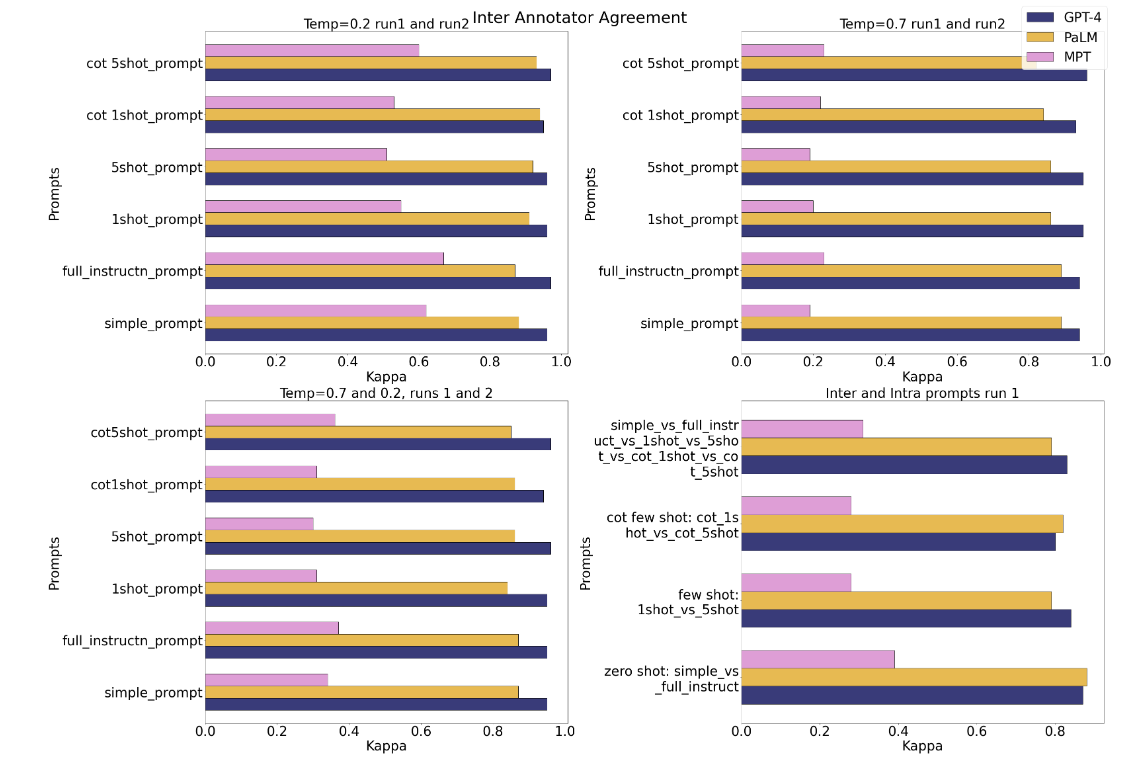}
\caption{Plots from Inter Annotator Agreement scores}
\label{fig:LLM_IAA}
\end{figure*}

\clearpage
\begin{table}[htb]
\subsection{Error Analysis}
\centering
\resizebox{\textwidth}{!}{
\begin{tabular}{c|l|c|l}
\hline
\textbf{Entity Pair} & \multicolumn{1}{c|}{\textbf{\begin{tabular}[c]{@{}c@{}}Scenario 1: \\ Crowd workers = Wrong Answer,  LLMs = Correct Answer\end{tabular}}} & \textbf{\begin{tabular}[c]{@{}c@{}}Scenario 2: \\ Crowd workers = Wrong Answer, LLMs = Wrong Answer\end{tabular}} & \multicolumn{1}{c}{\textbf{\begin{tabular}[c]{@{}c@{}}Scenario 3: \\ Crowd workers = Correct Answer, LLMs = Wrong Answer\end{tabular}}} \\ \hline
ORG-GPE & \begin{tabular}[c]{@{}l@{}}\textcolor{red}{\underline{Properties Atlas Financial Holdings}} , Inc. \\ corporate headquarters is located \\ at 150 Northwest Point Boulevard , \\ \textcolor{blue}{\underline{Elk Grove Village}} , Illinois 60007 , USA.\\ \\ Expert Label:\textsc{Headquartered in}\\ Crowd worker Label: \textsc{Operations in}\\ LLMs Label:\textsc{Headquartered in}\end{tabular} & \multicolumn{1}{l|}{\begin{tabular}[c]{@{}l@{}}Our \textcolor{red}{\underline{eWellness Corporate Office}} is located in \textcolor{blue}{\underline{Culver City}}\\  , California . eWELLNESS\\ \\ \\ Expert Label: \textsc{Operations in}\\ Crowd worker Label: \textsc{Formed in}\\ LLMs Label: \textsc{Headquartered in}\end{tabular}} & \begin{tabular}[c]{@{}l@{}}This Settlement Agreement ( " Agreement " ) is made effective \\ this 20th day of May , 2015 by and between \\ ActiveCare , Inc , a \textcolor{blue}{\underline{Delaware}} corporation ( the " Company " )  \\  , and \textcolor{red}{\underline{Advance Technology Investors , LLC}} ( " ATI " ) .\\ \\ \\ Expert Label:\textsc{Operations in}\\ Crowd worker Label: \textsc{Operations in}\\ LLMs Label: \textsc{No/Other Relation, Formed in}\end{tabular} \\ \hline

ORG-ORG & \begin{tabular}[c]{@{}l@{}}Michael D. Huddy , President / CEO and Director , joined \\ \textcolor{blue}{\underline{INTERNATIONAL BARRIER TECHNOLOGY INC}} in \\ February 1993 as President of the newly - formed\\  US Subsidiary, \textcolor{red}{\underline{Barrier Technology Corporation}} .\\ \\ \\ Expert Label:\textsc{Subsidiary of}\\ Crowd worker Label: \textsc{No/Other Relation, Shares of}\\ LLMs Label: \textsc{Subsidiary of}\end{tabular} & \multicolumn{1}{l|}{\begin{tabular}[c]{@{}l@{}}Our \textcolor{red}{\underline{Hawaii Gas}} entered into licensing agreements with \\ Utility Service Partners , Inc. and America's Water Heater\\  Rentals , LLC , both indirect subsidiaries of\\ \textcolor{blue}{\underline{Macquarie Group Limited}}, to enable these entities \\ to offer products and services to Hawaii Gas's customer base\\ \\ Expert Label:\textsc{Subsidiary of}\\ Crowd worker Label: \textsc{No/Other Relation, Subsidiary of, Shares of} \\ LLMs Label: \textsc{Agreement with}\end{tabular}} & \begin{tabular}[c]{@{}l@{}}On December 10 , 2014 , \textcolor{red}{\underline{Orbital Tracking Corp.}} purchased\\  certain contracts from Global Telesat Corp ,a Virginia corporation ( GTC ) \\ for \$ 250,000 pursuant to an asset purchase agreement by and among\\ Orbital Tracking Corp i, its wholly owned subsidiary Orbital Satcom, \\ \textcolor{blue}{\underline{GTC}} and World Surveillance Group , Inc. ( World ) , GTC's parent\\ \\ \\ Expert Label: \textsc{Subsidiary of}\\ Crowd worker Label: \textsc{Subsidiary of}\\ LLMs Label: \textsc{Agreement with}\end{tabular} \\ \hline

ORG-DATE & \begin{tabular}[c]{@{}l@{}}\textcolor{red}{\underline{Wishbone Pet Products Inc.}} was incorporated in the \\ State of Nevada on \textcolor{blue}{\underline{July 30 , 2009 .}}\\ \\ \\ \\ Expert Label: \textsc{Formed on}\\ Crowd worker Label: \textsc{No/Other Relation}\\ LLMs Label: \textsc{Formed on}\end{tabular} & None & \multicolumn{1}{c}{None} \\ \hline 

ORG-MONEY & \begin{tabular}[c]{@{}l@{}}\textcolor{red}{\underline{Personal Lines}} underwriting profit for the \\ three months ended September 30 , 2017\\ was \textcolor{blue}{\underline{\$40.8 million}} , compared to \$23.3 \\ million for the three months ended \\ September 30 ,2016 , an improvement \\ of \$17.5 million .\\ \\ \\ Expert Label: \textsc{Profit of}\\ Crowd worker Label: \textsc{No/Other Relation, Loss of}\\ LLMs Label: \textsc{Profit of}\end{tabular} & None & \multicolumn{1}{c}{None} \\ \hline

PERS-ORG & \begin{tabular}[c]{@{}l@{}}Mr. \textcolor{red}{\underline{Untermeyer}} also serves as senior program manager\\ with \textcolor{blue}{\underline{Southwest Research institute}} , San Antonio\\ \\ \\ Expert Label: \textsc{Employee of}\\ Crowd worker Label: \textsc{Founder of, Member of}\\ LLMs Label: \textsc{Employee of}\end{tabular} & \multicolumn{1}{l|}{\begin{tabular}[c]{@{}l@{}}Currently , Mr. \textcolor{red}{\underline{Morrison}} serves on the board of directors \\ of the \textcolor{blue}{\underline{Texas AM university}} , kingsville foundation \\ and the Rockport center for the arts.\\ \\ Expert Label: \textsc{Employee of}\\ Crowd worker Label: \textsc{Founder of}, \textsc{Member of}\\ LLMs Label: \textsc{Member of}\end{tabular}} & \begin{tabular}[c]{@{}l@{}}From September 2012 through June 2015 , Mr. \textcolor{red}{\underline{Kimmel}} has also \\ served on the board of directors of Electronic Magnetic Power \\ Solutions , which implements disruptive patented technology \\ licensed from \textcolor{blue}{\underline{Virginia Tech University}} for the express purpose\\ of alternative energy use in the consumer space .\\ \\ \\ Expert Label: \textsc{Employee of}\\ Crowd worker Label: \textsc{Employee of}\\ LLMs Label: \textsc{Member of}, \textsc{No/Other Relation}\end{tabular} \\ \hline

PERS-TITLE & \begin{tabular}[c]{@{}l@{}}Information regarding \textcolor{red}{\underline{Harel Gadot}} , Microbot Medical Inc.\\ \textcolor{blue}{\underline{Chairman}} , President and Chief Executive Officer , is set\\ forth above under Board of Directors .\\ \\ \\ Expert Label:\textsc{Title}\\ Crowd worker Label: \textsc{No/Other Relation}\\ LLMs Label:\textsc{Title}\end{tabular} & None & \begin{tabular}[c]{@{}l@{}}\textcolor{red}{\underline{Yvonne}} should contact her manager , \\ segment or region \textcolor{blue}{\underline{leader}} , or FTI Consulting s \\ Chief Ethics and Compliance Officer to discuss the gift .\\ \\ \\ Expert Label: \textsc{Title}\\ Crowd worker Label: \textsc{Title}\\ LLMs Label: \textsc{No/Other Relation}\end{tabular} \\ \hline
\end{tabular}}
\vspace{0.3cm}
\parbox{0.95\textwidth}{
  \caption{Qualitative Examples from our Error Analysis depicting the 3 prominent scenarios of how MTurk Crowd workers and LLMs demonstrated high confidence on answer choice}
  \label{tab:scenarios}
}
\end{table} 

\begin{figure*}[htb]
\subsection{Confusion Matrix for GPT-4}
\centering
\includegraphics[width=\textwidth]{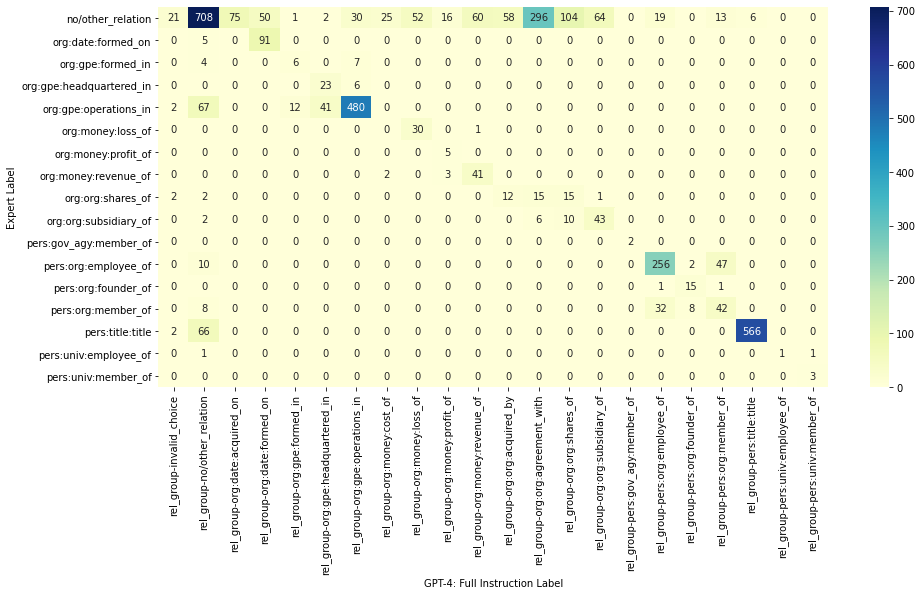}
\vspace{0.2cm} 
\parbox{0.95\textwidth}{
  \caption{Confusion Matrix for GPT-4 Zero Shot Prompt}
  \label{fig:zs-cm-gpt}
}
\end{figure*}

\begin{figure*}[h!]
\centering
\includegraphics[width=\textwidth]{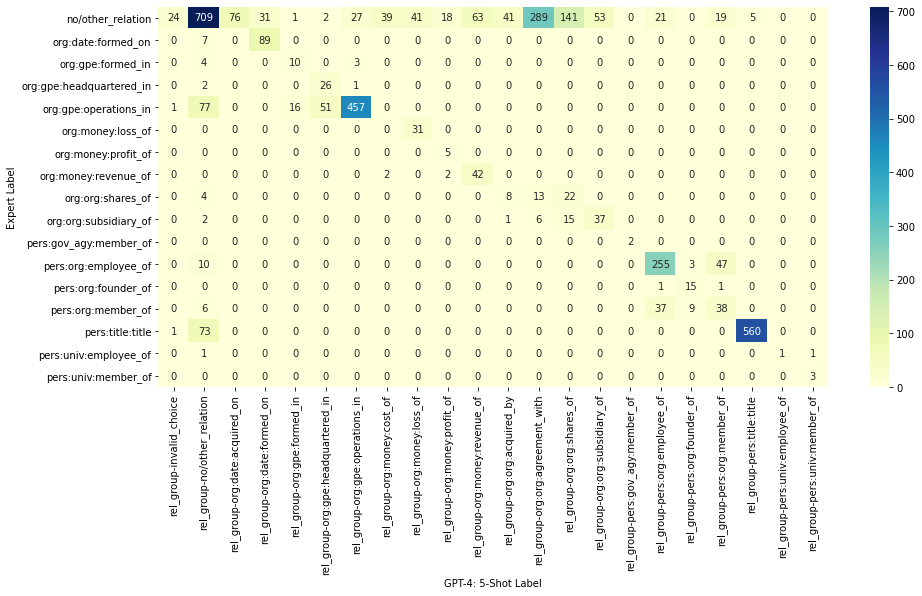}
\vspace{0.2cm}
\parbox{0.95\textwidth}{
  \caption{Confusion Matrix for GPT-4 Few Shot Prompt}
  \label{fig:fs-cm-gpt}
}
\end{figure*}

\begin{figure*}[hbt!]
\centering
\includegraphics[width=\textwidth]{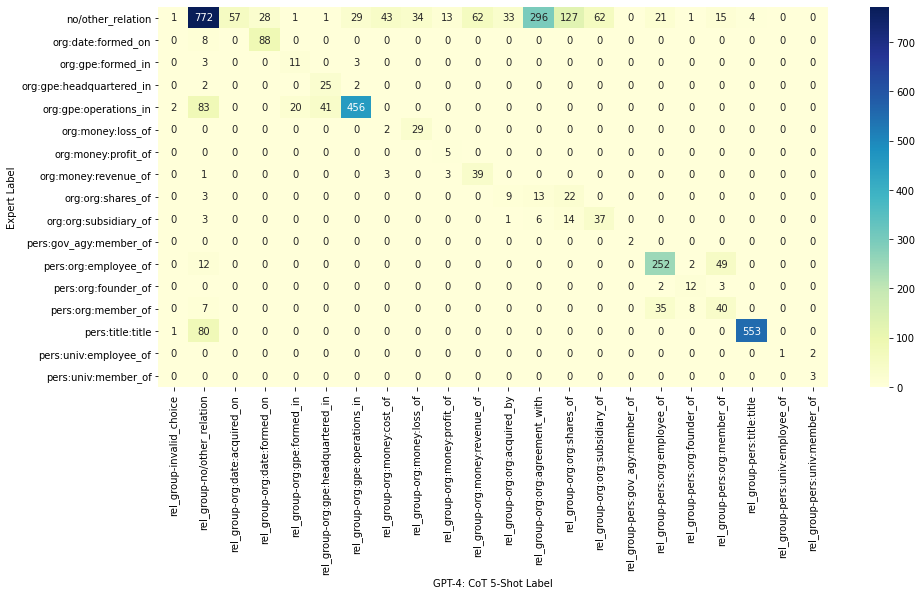}
\vspace{0.2cm}
\parbox{0.95\textwidth}{
  \caption{Confusion Matrix for GPT-4 Few Shot CoT Prompt}
  \label{fig:cot-cm-gpt}
}
\end{figure*}
\end{document}